\documentclass[10pt,a4paper]{article}
\usepackage[T1]{fontenc}
\usepackage[margin=1in,includehead,headheight=11pt,headsep=11pt,columnsep=2pc]{geometry}
\usepackage{mathptmx}
\usepackage{amsmath,amssymb}
\usepackage{microtype}
\usepackage{graphicx}
\usepackage{float}
\usepackage{caption}
\usepackage{fancyhdr}
\usepackage{enumitem}
\usepackage{cite}
\usepackage{xcolor}
\usepackage{url}
\usepackage{listings}
\usepackage[most]{tcolorbox}
\usepackage[colorlinks=true,linkcolor=blue,citecolor=blue,urlcolor=blue]{hyperref}
\hypersetup{
 pdftitle={Functionalizing AI-Generated 3D Shapes via Interior Geometry Construction},
 pdfauthor={Haoyu Tang, Fancheng Gao, Kan Chen, Peng Song},
 pdfsubject={Interior geometry construction for AI-generated 3D shapes},
 pdfkeywords={Shape modeling, interior geometry, generalized cylinders, vision-language models}
}
\setlist{leftmargin=1.5em,labelsep=0.5em,itemsep=0pt plus 0.1pt,
 parsep=0pt,topsep=0pt plus 1pt minus 0.5pt,partopsep=0pt}
\makeatletter

\let\supplementseccntformat\@seccntformat
\renewcommand{\section}{\@startsection{section}{1}{\z@}
 {-11pt plus -5.5pt minus -1pt}{0.001pt}
 {\raggedright\normalfont\normalsize\bfseries}}
\renewcommand{\subsection}{\@startsection{subsection}{2}{\z@}
 {-11pt plus -5.5pt minus -1pt}{0.001pt}
 {\raggedright\normalfont\normalsize\bfseries}}
\renewcommand{\subsubsection}{\@startsection{subsubsection}{3}{\z@}
 {-5.5pt plus -5.5pt minus -1pt}{0.001pt}
 {\raggedright\normalfont\normalsize\bfseries}}
\renewcommand{\paragraph}{\@startsection{paragraph}{4}{\z@}
 {5.5pt plus 2.25pt minus 1pt}{-0.5em}{\normalfont\normalsize\bfseries}}
\renewcommand{\@seccntformat}[1]{\csname the#1\endcsname.\hspace{0.5em}}
\renewcommand{\@maketitle}{%
 \newpage\null\vskip 20pt
 \begingroup
 \centering\setlength{\parskip}{0pt}
 {\fontsize{17.28}{20}\selectfont\bfseries\@title\par}
 \vskip 25pt
 {\normalsize\begin{tabular}[t]{c}\@author\end{tabular}\par}
 \vskip 18pt
 \endgroup
}
\makeatother

\makeatletter

\renewcommand{\@biblabel}[1]{[#1]\hspace{0.5em}}
\makeatother

\renewenvironment{abstract}{%
 \par\vspace{2pt}\hrule height 0.25pt\vspace{3pt}
 \begin{list}{}{\setlength{\leftmargin}{2pc}\setlength{\rightmargin}{2pc}
  \setlength{\topsep}{0pt}\setlength{\partopsep}{0pt}\setlength{\parsep}{0pt}}
 \item[]\normalsize\textbf{Abstract}\par\itshape\noindent
}{\end{list}\vspace{3pt}\hrule height 0.25pt\par}
\title{Functionalizing AI-Generated 3D Shapes\\via Interior Geometry Construction}
\author{Haoyu Tang$^1$\quad Fancheng Gao$^1$\quad Kan Chen$^2$\quad Peng Song$^1$\thanks{Corresponding author.}\\[5pt]
  \small $^1$Singapore University of Technology and Design\\
  \small $^2$Singapore Institute of Technology}
\date{}
\begin{document}
\maketitle
\thispagestyle{empty}
\begingroup
\setlength{\intextsep}{0pt}
\begin{figure}[H]
  \centering
  \includegraphics[width=0.96\linewidth]{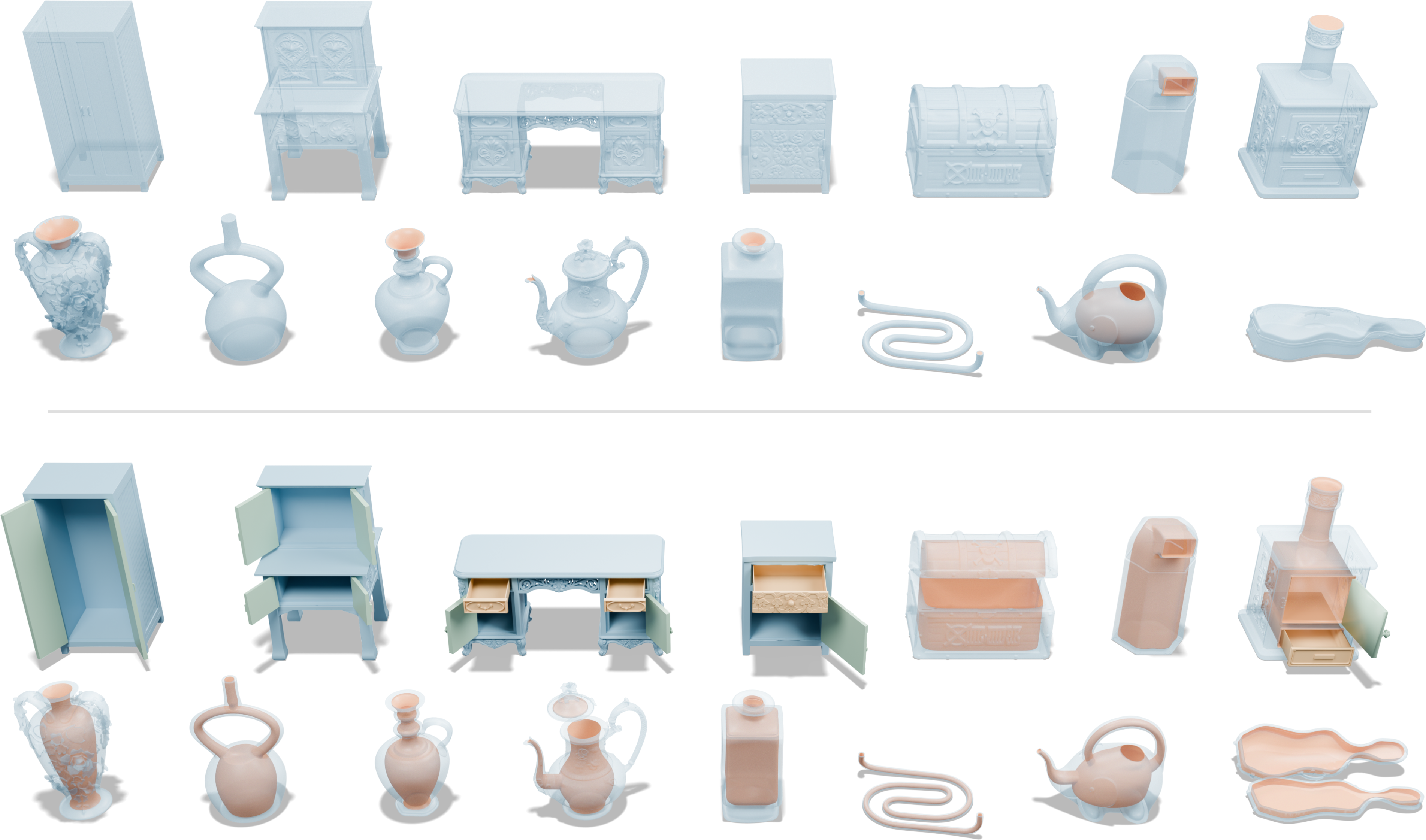}
  \caption{We propose a computational approach to automatically construct interior geometry for (top) AI-generated 3D shapes, such that (bottom) the processed 3D shapes are functional and can be used for downstream applications such as simulation and fabrication.}
  \label{fig:teaser}
\end{figure}
\endgroup

\begin{abstract}
AI-based 3D generation models such as Hunyuan 3D and Meshy can produce visually appealing 3D shapes from an image or text prompt. 
However, they typically focus on an object's external appearance, often producing incomplete or implausible interior geometry.
Consequently, the generated shapes are not guaranteed to be functionally valid, limiting their direct use in downstream applications such as simulation and fabrication.
In this paper, we present a computational approach to functionalize AI-generated 3D shapes through interior geometry construction. 
Our key observations are that a part’s functional role defines its interior geometry, and such geometry is often smooth, relatively simple, and closely related to the part’s external shape.
Inspired by these observations, we introduce an {\em interior-aware part graph} to encode information necessary for constructing interior geometry within an  AI-generated 3D shape, and build the graph via VLM-based  shape segmentation and graph attribute inference.
Guided by the graph, we model the interior geometry of each part as a generalized cylinder and infer its parameters from the part's external geometry through an optimization.
The resulting 3D shapes are functionally valid while preserving their AI-generated external appearance.
We demonstrate our approach on a variety of AI-generated 3D shapes, including vessels, containers, and storage furniture, and fabricate selected examples for functional validation.

\end{abstract}  
\twocolumn

\section{Introduction}
\label{sec:intro}

\begin{figure}[!t]
	\centering
	\includegraphics[width=\linewidth]{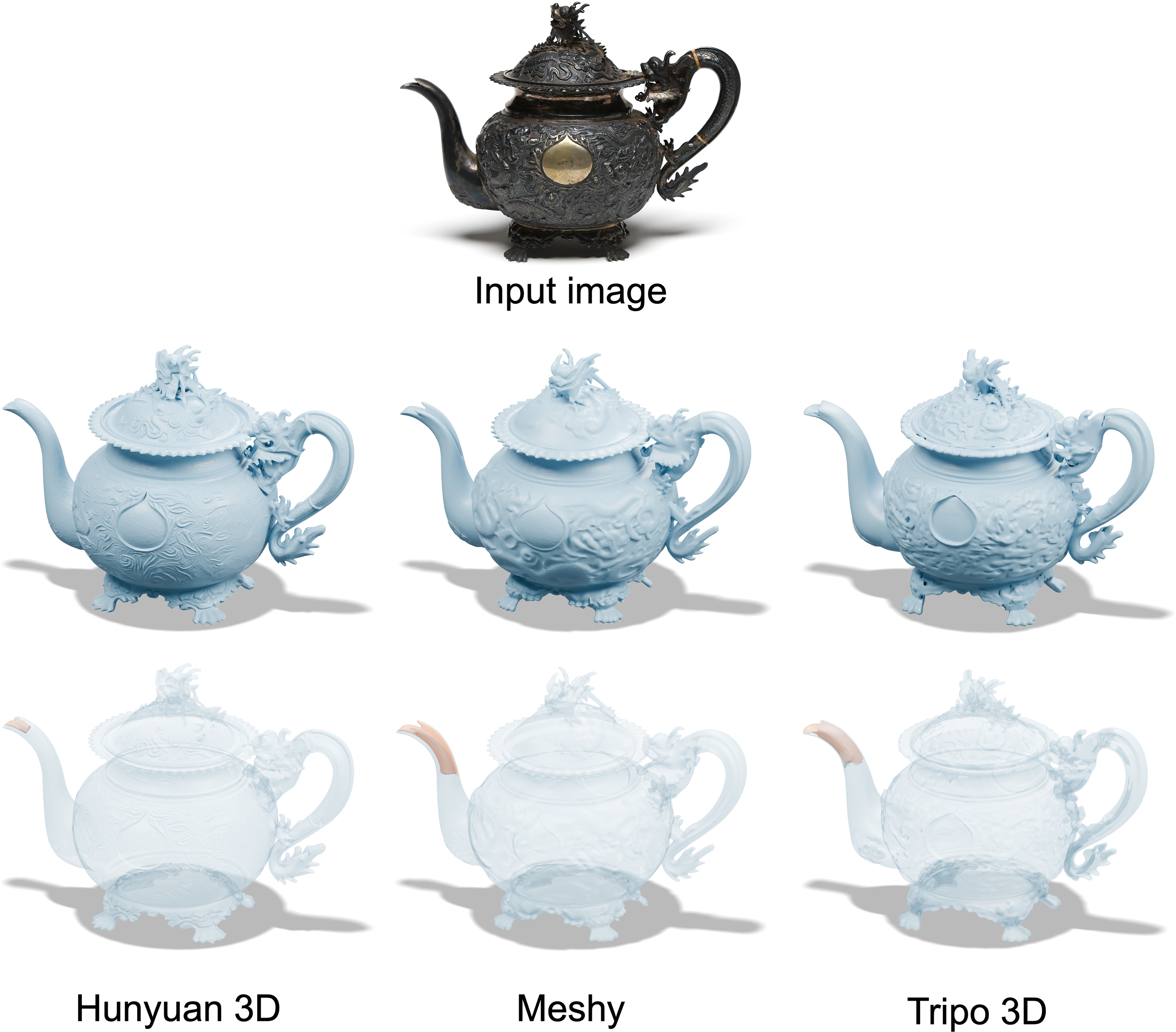}
	\caption{AI-based 3D generation primarily focuses on (middle) generating an object's external appearance that matches (top) user-specified input such as an image prompt, (bottom) producing incomplete or implausible interior geometry.}
	\label{fig:motivation}
\end{figure}

Recent advances in AI-based 3D generation have enabled the creation of visually compelling 3D shapes from lightweight user inputs, substantially reducing the effort required compared with conventional 3D modeling.
In particular, AI-based 3D generation models, such as Hunyuan 3D~\cite{2026-Hunyuan3D}, Meshy~\cite{2026-Meshy}, and Tripo 3D~\cite{2026-Tripo3D}, can generate detailed 3D shapes that closely match objects described by an image or text prompt within seconds or minutes; see Figure~\ref{fig:motivation} (middle).
Despite this progress, these AI models primarily focus on generating an object's external appearance, without explicitly reasoning about the object's functionality.
Consequently, AI-generated 3D shapes often either lack meaningful interior geometry or contain interior geometry that is inconsistent with their intended functions; see Figure~\ref{fig:motivation} (bottom).
While such 3D shapes may suffice for visualization, they are often unsuitable for direct use in downstream applications that require functionally valid geometry, such as physically-based simulation, digital fabrication, and robotic interaction.

In this work, we aim to enable the creation of 3D shapes that are {\em not only visually appealing but also functionally plausible} from lightweight user inputs (e.g., images or text prompts). 
To achieve this goal, we address the key research problem of {\em functionalizing AI-generated 3D shapes of everyday objects through interior geometry construction.}
However, this problem is non-trivial due to two reasons. 
First, an object may consist of parts with geometrically and functionally distinct interior geometry. 
Consider a teapot: its body requires a hollow cavity, its spout requires a tubular channel connected to the cavity, while its handle should remain solid. 
Therefore, interior construction cannot be achieved by simply shelling or hollowing the entire shape; different parts require different geometric treatments and appropriate connections between them. 
Second, everyday objects with functional interiors exhibit considerable structural diversity. 
For example, a tube, a teapot, and a cabinet all contain interior spaces, yet they differ substantially in their functions, interior structures, and  part organizations. 

To address these challenges, we make two key observations.
First, interior geometry construction becomes considerably more manageable when performed at the level of functional parts.
Rather than constructing the entire interior as a monolithic geometry, we can decompose an object into functional parts, infer the functional role of each part, determine whether each part requires interior geometry, and construct such geometry when necessary.
Second, the interior geometry of individual parts is typically smooth and relatively simple, as interior spaces are commonly designed to contain, store, accommodate, or transfer physical entities.
Moreover, the interior geometry of a part is often closely related to its external geometry, which provides useful geometric cues for inferring the interior shape.

In this paper, we propose a computational approach to constructing interior geometry within AI-generated 3D shapes to make them functional.
In particular, we make two technical contributions:
\begin{itemize}	
	\item 
	Inspired by the first observation, we introduce an {\em interior-aware part graph} to represent functional parts of an AI-generated 3D shape and encode information necessary for constructing interior geometry within the parts,  and build this graph via VLM-based shape segmentation and graph attribute inference.
	
	\vspace*{1mm}
	\item 
	Inspired by the second observation, we model the interior geometry of each part as a generalized cylinder, and infer its parameters from the part's external geometry through an optimization, guided by the interior-aware part graph. 
\end{itemize}

Our approach ensures that the processed 3D shape achieves the intended functionality through the constructed interior geometry while preserving the AI-generated external appearance.
We evaluate our approach on diverse everyday objects generated by state-of-the-art AI models that are widely used for 3D generation; see Figure~\ref{fig:teaser}  for examples.

\begin{figure*}[tbp]
	\centering
	\includegraphics[width=0.99\linewidth]{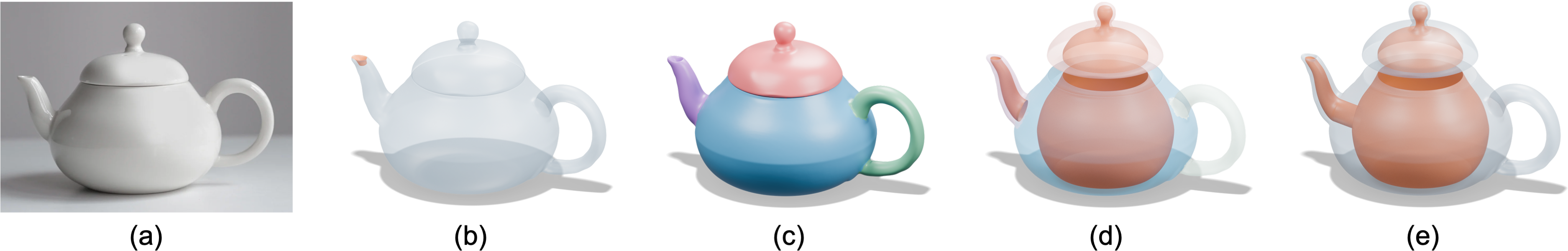}
	\vspace*{-1mm}
	\caption{Overview of our approach. 
		Given (a) an image prompt, we use appearance-driven AI models such as Hunyuan 3D to (b) generate a 3D shape that matches the prompt.
		To functionalize the 3D shape, we (c) segment the input shape into functional parts and represent these parts and their relations via an interior-aware part graph.
		Guided by the graph, we (d) construct interior geometry for each identified part and compose the parts' interior geometries together with the AI-generated exterior surface to form (e) the output shape. }
	\label{fig:overview}
	\vspace*{-2mm}
\end{figure*}

\section{Related Work}

We review and classify existing works on AI-based 3D shape generation and discuss their relation to our work.

\noindent
\textbf{Appearance-driven 3D shape generation} 
aims to generate 3D shapes that match the visual appearance of objects depicted in input images or described by text prompts.
Approaches for appearance-driven 3D shape generation can be broadly classified into three categories. 
First, prior-guided 3D optimization approaches, such as DreamFusion~\cite{Poole-2023-DreamFusion}, Magic3D~\cite{Lin-2023-Magic3D}, ProlificDreamer~\cite{Wang-2023-ProlificDreamer}, and DreamGaussian~\cite{Tang-2024-DreamGaussian}, leverage pretrained 2D diffusion models to optimize a 3D representation whose rendered views align with input text or images. 
Second, multi-view generation approaches first synthesize consistent views of an object and then reconstruct its 3D geometry from these views, as demonstrated by Zero-1-to-3~\cite{Liu-2023-Zero123}, MVDream~\cite{Shi-2023-MVDream}, Wonder3D~\cite{Long-2023-Wonder3D}, and InstantMesh~\cite{Xu-2024-InstantMesh}. 
Third, more recently, feed-forward 3D generation approaches learn a direct mapping from text or images to 3D representations, avoiding costly per-instance optimization and enabling efficient asset creation. 
Representative works include LRM~\cite{Hong-2023-LRM}, Michelangelo~\cite{Zhao-2023-Michelangelo}, TripoSR~\cite{Tochilkin-2024-TripoSR}, CLAY~\cite{Zhang-2024-CLAY}, Direct3D~\cite{Wu-2024-Direct3D},  TRELLIS~\cite{Xiang-2024-TRELLIS}, Hunyuan 3D~\cite{Zhao-2025-Hunyuan3D}, and Meshy~\cite{Xu-2026-Meshy}. 

Despite their different generation paradigms, these methods primarily emphasize visual fidelity and input alignment, and generally focus on exterior appearance rather than functionality.
In this work, we take the 3D shapes generated by the above approaches, in particular the feed-forward approaches such as Hunyuan 3D~\cite{2026-Hunyuan3D}, and propose a computational approach that automatically creates interior geometry within these AI-generated 3D shapes to make them functional.

\noindent
\textbf{Agentic 3D shape generation.} \
Recent advances in Large Language Models (LLMs) and Vision-Language Models (VLMs) have enabled agentic 3D shape generation, where AI agents turn 3D modeling into an iterative programming and reasoning task rather than a one-shot prediction.
Existing agentic 3D shape generation approaches can be broadly classified into three categories.
First, agentic procedural approaches, such as 3D-GPT~\cite{Sun-2025-3D-GPT}, SceneX~\cite{Zhou-2025-SceneX}, and LandCraft~\cite{Liu-2026-LandCraft}, translate user inputs into structured parameters or procedural representations for shape construction.
Second, code-based approaches formulate shape generation as code synthesis, producing executable CAD code or programs that explicitly construct geometry, as demonstrated by ShapeCraft~\cite{Zhang-2025-ShapeCraft}, CADCrafter~\cite{Chen-2025-CADCrafter}, LLM4CAD~\cite{Li-2025-LLM4CAD}, and multi-agent CAD code generation~\cite{Liu-2026-MultiAgentCAD}.
Third, feedback-driven approaches further introduce a perception--action loop, in which agents inspect intermediate modeling results and iteratively refine the generated geometry, as demonstrated by BlenderAlchemy~\cite{Huang-2024-BlenderAlchemy}, SceneCraft~\cite{Hu-2024-SceneCraft}, and CAD-Assistant~\cite{Mallis-2025-CADAssistant}.
General-purpose AI models, such as GPT-6 Astra~\cite{2026-GPT6Astra} and Claude~\cite{2026-Claude}, increasingly follow this paradigm by interacting with 3D modeling tools such as Blender to construct, inspect, and iteratively modify 3D shapes to match users' image and/or text prompts.

These agentic approaches are powerful in generating 3D shapes that can be procedurally modeled or represented by code such as cities, buildings, and CAD models.  
However, when the target shape (e.g., depicted by an input image) consists of freeform geometries, these approaches can only approximate rather than reproducing them as above appearance-driven 3D shape generation does.
We have shown that 3D shapes functionalized by our approach can better preserve appearance in the image prompt than those produced by agentic approaches such as  GPT-6 Astra~\cite{2026-GPT6Astra}; see Figure~\ref{fig:result_shape}.

\noindent
\textbf{Physics-aware 3D shape generation} 
incorporates physics-related constraints in the 3D shape generation process, aiming to make use of the generated 3D shapes in the physical world.
In particular,
fabricability~\cite{Faruqi-2023-Style2Fab, Faruqi-2024-ShapingRealities},
self-supporting under gravity~\cite{Dang-2026-DensiCrafter, Chen-2024-Atlas3D, Guo-2024-PhysicallyCompatible}, and
structural soundness~\cite{Xu-2025-FunctionsBetter, Faruqi-2025-MechStyle, Zhan-2025-PhysiOpt} 
have been incorporated when generating monolithic 3D shapes while 
assemblability~\cite{Pun-2025-BrickStructure, Ge-2025-LEGOMaker, Xu-2025-LegoACE, Botzer-2026-Laserify} and 
articulation~\cite{Chen-2025-FreeArt3D, Lu-2025-DreamArt, Le-2025-ArticulateAnything, Chen-2025-ArtiLatent}
have been incorporated when generating assemblies consisting of multiple parts.
Recently, researchers have become interested in incorporating functionality when generating 3D shapes, including
designing functional objects for fitting human/character bodies~\cite{Guo-2025-CRAFT}, 
generating functional artifacts for reconfiguring everyday environments~\cite{Arabi-2026-EUREXA}, and
functionalizing furniture via adding missing parts such as connectors~\cite{Zhao-2026-Functionalization}.

Following this line of research, our work aims to generate everyday objects with functionality that relies on their interior space.
To this end, we leverage existing appearance-driven AI models to generate visually appealing 3D shapes and then functionalize these AI-generated 3D shapes through interior geometry construction.

\section{Problem Formulation}
\label{sec:problem}

In this paper, we focus on everyday objects whose functionality depends on their interior space, such as containing, storing, accommodating, and transferring.
Typical examples are vessels, containers, storage furniture, bins, and pipes; see Figure~\ref{fig:teaser}. 

\begin{figure*}[t]
	\centering
	\includegraphics[width=\linewidth]{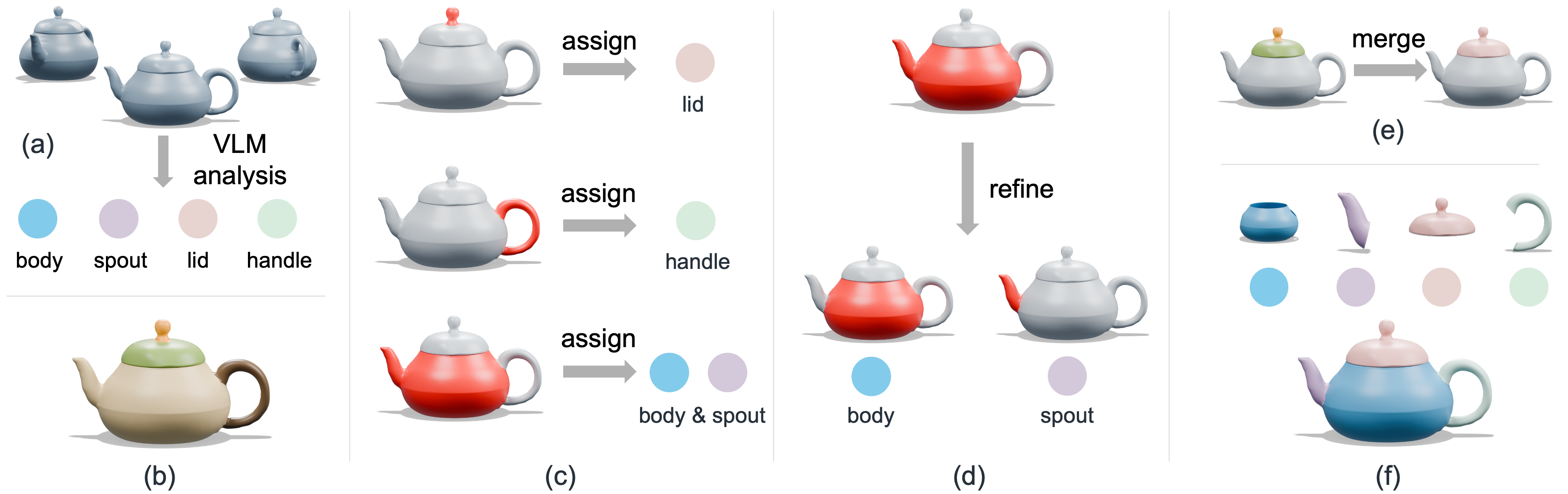}
	\caption{
		VLM-guided shape segmentation. 
		(a) Analyze multiple views of a teapot using a VLM, leading to an inventory of four functional components (i.e., body, spout, lid, and handle). 
		(b) Segment the teapot into four parts using a learning-based method.
		(c) Assign each segmented part (rendered in red) to one or multiple functional components. 
		(d) Refine the body--spout part by splitting it into two parts (i.e., a body and a spout), corresponding to the body and spout components.
		(e) Merge the knob part and lid base part into a single part, corresponding to the lid component.
		(f) The segmented teapot consists of four functional parts.
	}
	\vspace*{-3mm}
	\label{fig:shape_segmentation}
\end{figure*}

\noindent
\textbf{Input shapes.} \
We leverage state-of-the-art appearance-driven AI models such as Hunyuan 3D~\cite{2026-Hunyuan3D}, Meshy~\cite{2026-Meshy}, and Tripo 3D~\cite{2026-Tripo3D} to generate 3D shapes of the above everyday objects from users' image or text prompt.
These 3D shapes are visually appealing and match the prompt.
However, they usually contain missing or incorrect interior geometry, rendering them non-functional in simulation and the real world. 
Our approach takes the AI-generated 3D shape as the input; see Figure~\ref{fig:overview} (a\&b).
To facilitate processing, we remove AI-generated interior geometry (if any) via extracting the shape's outer hull~\cite{Zhou-2016-MeshArrangements}, where geometries that are inaccessible from the shape's exterior surface are discarded.

\noindent
\textbf{Shape processing.} \
Given an AI-generated 3D shape, our goal is to process the 3D shape to make it functional while preserving the AI-generated external appearance.
Therefore, we have two requirements on the shape processing.
\begin{enumerate}
    \item 
    {\em Functional plausibility.} \
    It should construct interior geometry within the input shape that is necessary to make the  shape functional such as cavities and channels.
  
     \vspace*{1mm}
    \item 
    {\em Appearance preservation.} \
    It should preserve the AI-generated external appearance such that the processed shape still matches the user-specified prompt.
\end{enumerate}

\noindent
\textbf{Overview of our approach.} \
We propose a two-step approach to address the problem; see Figure~\ref{fig:overview} for an overview. 
In the first step, we understand functional parts and their relations in an AI-generated 3D shape that support the shape's functionality, and encode the information into an interior-aware part graph; see Figure~\ref{fig:overview}(c) and Section~\ref{sec:shape_understanding}.
In the second step, we leverage the part graph to guide the interior geometry construction while preserving the AI-generated appearance.
In particular, we first construct the interior geometry of each identified part as a generalized cylinder and then compose the parts' interior geometries together with the AI-generated exterior surface to form the output shape; see Figure~\ref{fig:overview}(d\&e) and Section~\ref{sec:construction}.

\section{Shape Understanding}
\label{sec:shape_understanding}

Given an AI-generated 3D shape, we segment the shape into functional parts, reason about how each part's interior geometry is defined to support the shape's functionality, and encode all these information into an interior-aware part graph.

\subsection{VLM-guided Shape Segmentation}
\label{subsec:shape_segmentation}

We perform shape segmentation to identify functional parts in the AI-generated 3D shape, represented by a polygon mesh $\mathcal{M}$.
Traditional geometry-based methods are not well suited to this task, as they segment 3D shapes primarily based on geometric properties without explicitly considering part semantics. 
Therefore, we adopt a learning-based method for our segmentation task. 
To ensure the segmented parts correspond to functional components, we use a VLM to identify functional components present in the shape for guiding the segmentation process.
Specifically, our VLM-guided shape segmentation consists of five steps; see Section~S1 of the supplementary material for details.

\begin{enumerate}

\item
{\em VLM-based shape analysis.} \
We use a VLM, such as Qwen~\cite{Qwen-API}, to analyze multi-view renderings of the input mesh $\mathcal{M}$ and form an inventory of visible functional components $\{ C_i \}$; see Figure~\ref{fig:shape_segmentation}(a).
For each component $C_i$, the inventory records a semantic description $\ell_i$ that identifies its function.
When multiple components share the same category (e.g., two handles of a vase), their descriptions also specify their positions relative to other components to distinguish individual instances.

\vspace*{1.0mm}
\item
{\em Initial shape segmentation.} \
We apply a learning-based method P3-SAM~\cite{Ma-2026-P3-SAM} to segment mesh $\mathcal{M}$ to obtain an initial set of parts denoted by $\{ P_j \}$.
Note that each initial part may not coincide with a functional component:
a part may correspond to a complete component, a fragment of a component, or portions of multiple components;  see Figure~\ref{fig:shape_segmentation}(b) for an example.

\vspace*{1.0mm}
\item
{\em Part assignment.} \
For each initial part $P_j$, the VLM receives the inventory and multi-view renderings highlighting $P_j$, with the remaining mesh shown in gray as context; see Figure~\ref{fig:shape_segmentation}(c).
Using the component descriptions $\ell_i$, the VLM determines whether $P_j$ belongs to a single component or spans multiple components. 
A part corresponding to one component $C_i$ or its fragment receives an \emph{assigned} status while a part spanning multiple components receives a \emph{mixed} status. 
Figure~\ref{fig:shape_segmentation}(c) illustrates these cases: 
the knob fragment is assigned to the lid component,
the complete handle part is assigned to the handle component, and 
the body--spout part is marked as \emph{mixed}.

\vspace*{1.0mm}
\item
{\em Part refinement.} \
Parts with a \emph{mixed} status are submitted to P3-SAM for refining the segmentation.
The newly segmented parts replace the original part and each of them undergoes the same part assignment step using the existing inventory, as illustrated in Figure~\ref{fig:shape_segmentation}(d).
We repeat this process until there is no part with a \emph{mixed} status in the segmentation result.

\vspace*{1.0mm}
\item
{\em Part merging.} \
In this step, we merge parts assigned to the same component.
Figure~\ref{fig:shape_segmentation}(e) shows the knob part and the lid base part are merged into a single part, corresponding to the lid component.
\end{enumerate}

Our approach terminates when every functional component $C_i$ identified by the VLM has a corresponding part $P_i$ in the segmentation.
Our approach does not guarantee that functional parts are always correctly segmented in any input shape.
For example, a lid may be mistaken for part of the body when there are no clear visual cues distinguishing them.
In our experiments, we found this approach works well for common everyday objects; see Figure~\ref{fig:result_shape} (right) for our segmentation results. 

\subsection{Interior-aware Part Graph}
\label{subsec:part_graph}

Given the functional parts identified in Section~\ref{subsec:shape_segmentation}, how interior geometry should be constructed within the parts remains unclear.
To this end, we propose an {\em interior-aware part graph} to represent information necessary for interior geometry construction, and infer the information using a VLM.

\noindent
\textbf{Interior-aware part graph.} \ 
We represent the segmented 3D shape as a part graph $G=(V,E)$, where each node represents a functional part and each edge connects two geometrically adjacent parts.
We augment the nodes and edges of the graph with attributes that are necessary for interior geometry construction.
We term this augmented part graph an {\em interior-aware part graph}.

\begin{figure}[tbp]
	\centering
	\includegraphics[width=\linewidth]{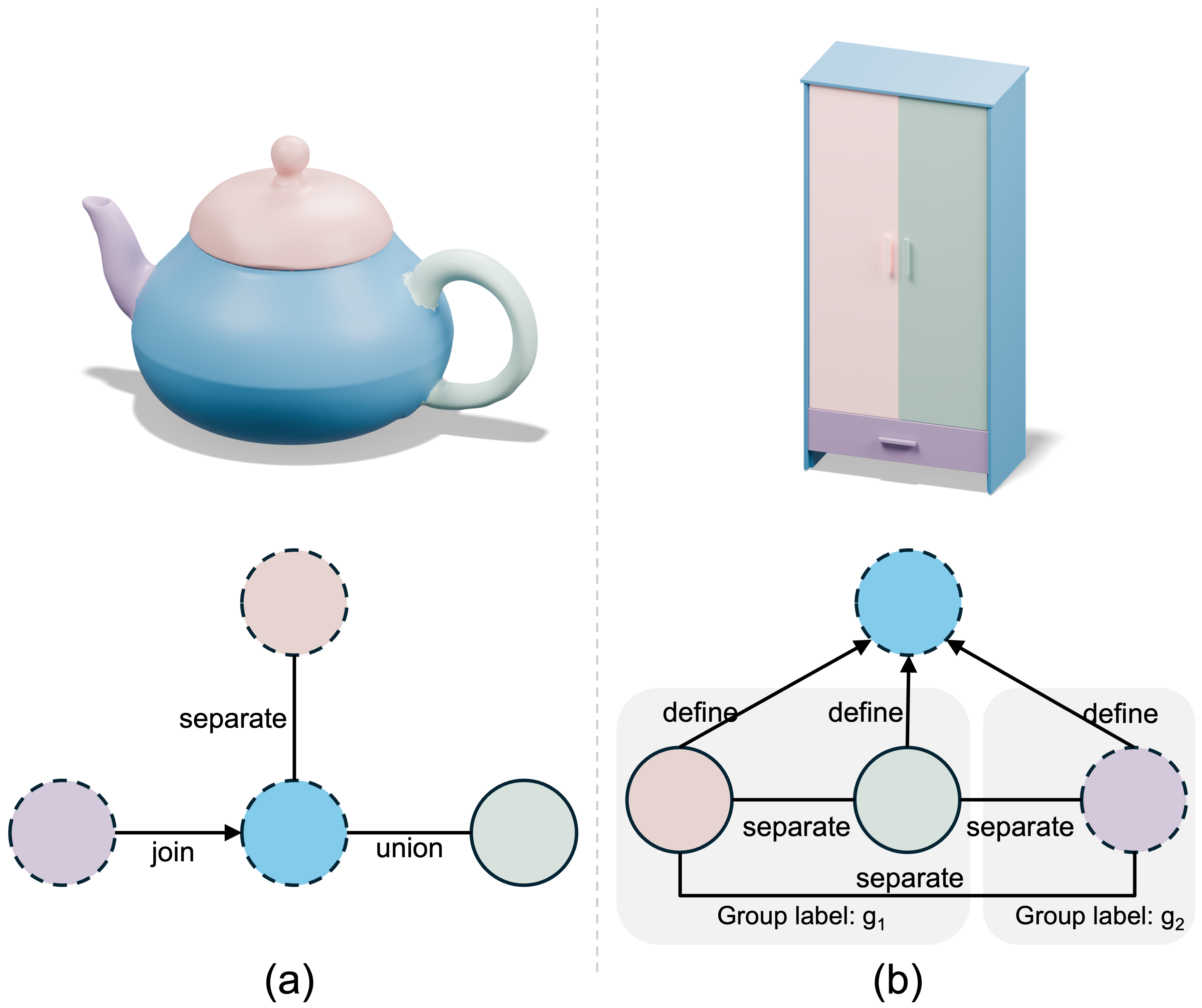}
	\caption{
		Two examples of interior-aware part graphs. 
		Node colors match the corresponding parts.  
		Dashed and solid node outlines indicate $h_i=1$ and $h_i=0$, respectively.
		(a) The teapot has three types of edges (i.e., $\mathit{join}$, $\mathit{union}$, and $\mathit{separate}$).
		(b) The wardrobe has two types of edges (i.e., $\mathit{define}$ and $\mathit{separate}$), and two groups of nodes (i.e., a group with two doors and a group with a drawer).
		Note that $\mathit{join}$ and $\mathit{define}$ edges are directed.
	}
	\label{fig:part_graph_construction}
\end{figure}

\noindent 
{\em Node attributes.} \
Each node $v_i\in V$ carries a binary cavity attribute $h_i\in\{0,1\}$, indicating whether the corresponding part should have its own interior space:
\begin{itemize}
    \item
    $h_i=1$. The part should have its own interior space, such as the body, spout, and lid of a teapot, and the body and drawer of a wardrobe; see Figure~\ref{fig:part_graph_construction}.
    
    \vspace*{1.0mm}
    \item 
    $h_i=0$. The part should not have its own interior space (i.e., the part is solid), such as the handle of a teapot, and the two doors of a wardrobe; see again Figure~\ref{fig:part_graph_construction}.
\end{itemize}

\noindent
{\em Edge attributes.} \
Each edge $e_k \in E$ carries an attribute $t_k$ specifying how the two associated parts, denoted by $v_i$ and $v_j$, are treated for interior geometry construction.
In particular, there are four types of edges:
$t_k \in \{\mathit{join}, \ \mathit{define},   \ \mathit{union}, \ \mathit{separate}  \}$.
\begin{itemize}
    \item 
    $\mathit{Join}$. \ 
    A directed edge from source part $v_i$ (with $h_i=1$) to target part $v_j$ (with $h_j=1$) that merges the parts and connects their interior spaces by extending the interior geometry of source part $v_i$.
    Figure~\ref{fig:part_graph_construction}(a) shows such an edge from the spout to the body of a teapot.
   
     \vspace*{1.0mm}
    \item 
    $\mathit{Define}$. \ 
    A directed edge from source part $v_i$ to target part $v_j$ (with $h_j=1$) in which the geometry of source part $v_i$ helps to define the interior space within the target part $v_j$.
    Figure~\ref{fig:part_graph_construction}(b) shows such edges from a door/drawer to the body of a wardrobe. 
   
    \vspace*{1.0mm}
    \item 
    $\mathit{Union}$. \
    An undirected edge that simply merges two associated parts $v_i$ and $v_j$.
    Figure~\ref{fig:part_graph_construction}(a) shows such an edge between the body and the handle of a teapot.
    
    \vspace*{1.0mm}
    \item 
    $\mathit{Separate}$. \
    An undirected edge that keeps two associated parts $v_i$ and $v_j$ as two separate parts.
    Figure~\ref{fig:part_graph_construction}(a) shows such an edge between the lid and the body of a teapot while 
    Figure~\ref{fig:part_graph_construction}(b) shows such edges between each door and the drawer of a wardrobe. 
\end{itemize}

For $\mathit{define}$ edges, we group source nodes that define the same interior space within a common target node, and assign each group a label $g_l$.
A group may contain a single source node or multiple source nodes.
In Figure~\ref{fig:part_graph_construction}(b), the two doors share a group since they together define the interior space within the body, while the drawer itself forms a group.

\noindent
\textbf{VLM-based graph attribute inference.} \
Given the segmented shape and the corresponding part graph, we use a VLM, such as Qwen~\cite{Qwen-API}, to infer the node and edge attributes including:
\begin{enumerate}
	\item 
	Node cavity attribute $h_i$ for each node $v_i$; 
	
	\vspace*{1.0mm}
	\item 
	Edge type attribute $t_k$ for each edge $e_k$;
	
	\vspace*{1.0mm}
	\item 
	Direction for each directed edge (with $\mathit{join}$ or $\mathit{define}$ type); and
	
	\vspace*{1.0mm}
	\item 
	Group label $g_l$ for source nodes of edges with $\mathit{define}$ type. 
\end{enumerate}

In our implementation, we render the shape with each part highlighted from multiple views.
We take all the rendered images as the input to the VLM, design a prompt that specifies the attribute definitions above, and  ask the VLM to
predict the values of the above attributes by reasoning each part's intended function.
The VLM returns all the attribute values in a single structured response.
Please refer to Section~S2 of the supplementary material for details.

\section{Shape Processing}
\label{sec:construction}

Guided by the interior-aware part graph, we process the AI-generated 3D shape to make it functional in this section. 
We first introduce a parametric representation for interior geometry to be constructed within each part (with $h_i=1$) in Section~\ref{sec:geometric_representation}.
Then, we present an optimization approach for constructing the interior geometry within a part based on the observed geometry and part graph in Section~\ref{sec:interior_geometry_construction}.
Lastly, we compose the constructed part interior geometries together with the AI-generated exterior surface to create a functional 3D shape in Section~\ref{sec:assembly}.

\begin{figure}[!t]
	\centering
	\includegraphics[width=0.8\linewidth]{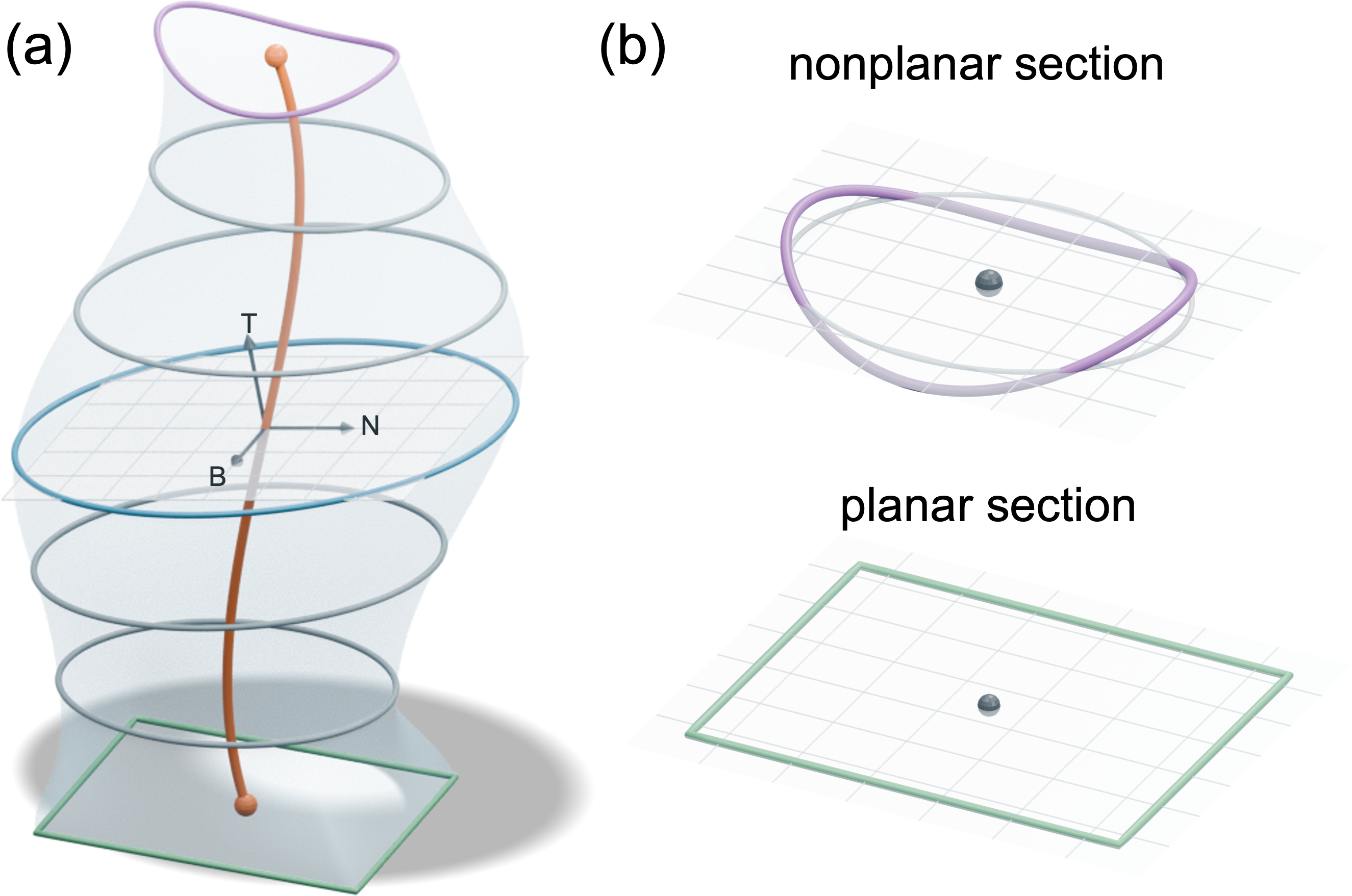}
	\caption{
		Generalized cylinder representation. 
		(a) A spine curve carries a field of closed sections;
		the blue section shows its local orthonormal frame $\mathbf{R}(u)$ and reference plane. 
		(b) A nonplanar section has varying $z$ in its local frame, while a planar section has $z=0$ in its local frame.}
	\vspace*{-2mm}
	\label{fig:geometric_representation}
\end{figure}

\subsection{Part Interior Geometry Representation}
\label{sec:geometric_representation}

The interior geometry of functional parts such as cavities, channels, and compartments is typically smooth and relatively simple, even when their external geometry is detailed.
Therefore, we choose {\em generalized cylinder (GC)}~\cite{Ballard-1982-ComputerVision} as a parametric representation for interior geometry to be constructed within a part.

A generalized cylinder is a parametric surface $\mathbf{s}(u,v)$ formed by sweeping a cross-sectional curve $\mathbf{q}(u,v)$ along a 3D spine curve $\mathbf{c}(u)$,
\begin{equation}
	\mathbf{s}(u,v)=\mathbf{c}(u)+\mathbf{R}(u)\mathbf{q}(u,v),
	\label{eq:gc_representation}
\end{equation}
where $u\in[0,1]$ is the longitudinal parameter and $v\in[0,1)$ is the periodic parameter along each section.
We represent the spine $\mathbf{c}(u)$ as a B-spline curve.
The matrix $\mathbf{R}(u)=[\mathbf{N}(u),\mathbf{B}(u),\mathbf{T}(u)]$ defines an orthonormal local frame with origin $\mathbf{c}(u)$.
The axes $\mathbf{N}$ and $\mathbf{B}$ span the local reference plane, while $\mathbf{T}$ is normal to it; see Figure~\ref{fig:geometric_representation}(a).
The section curve $\mathbf{q}(u,v)=(x,y,z)^{\mathsf T}$ is defined in the local frame.

The section curve $\mathbf{q}(u,v)$ should be closed and periodic in $v$ for each fixed $u$.
The curve's shape and dimensions may vary along the $u$ direction.
Moreover, the curve can be planar (by setting $z=0$) or nonplanar (by allowing nonzero $z$); see Figure~\ref{fig:geometric_representation}(b).
To this end, we represent the section curve $\mathbf{q}(u,v)$ using a Fourier series in $v$, with its coefficients modeled as B-spline functions of $u$.
Please refer to Section~S3 of the supplementary material for details about $\mathbf{q}(u,v)$.

\subsection{Part Interior Geometry Construction}
\label{sec:interior_geometry_construction}

The interior geometry of a part can be either defined by itself (e.g., body of a teapot) or defined by itself together with other parts (e.g., interior space of a wardrobe is defined by its body, two doors, and a drawer); see again Figure~\ref{fig:part_graph_construction}.
Therefore, we classify part interior geometry construction into two cases:
1) independent interior construction; and 
2) dependent interior construction.
For the second case, the dependency is specified by edges with a $\mathit{define}$  type (or simply, $\mathit{define}$ edges) in the part graph. 

\begin{figure}[tbp]
    \centering
    \includegraphics[width=\linewidth]{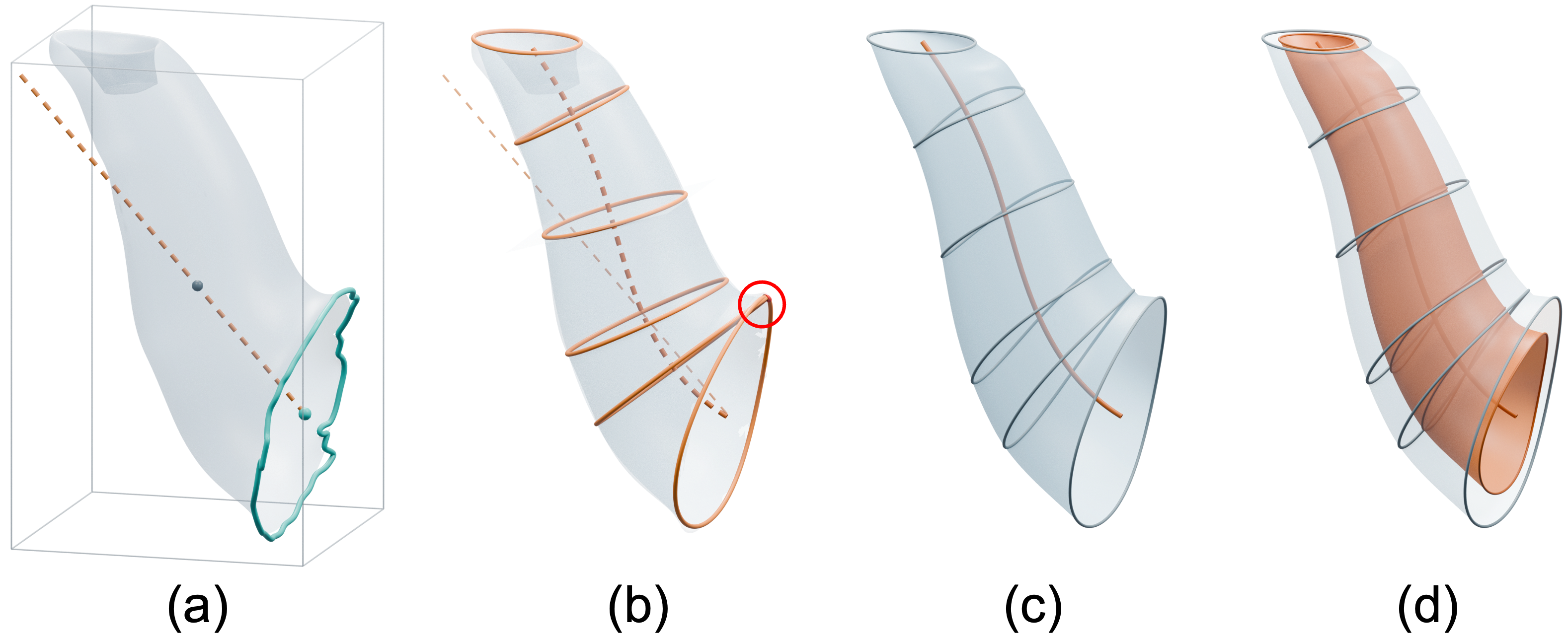}
    \caption{
    Independent interior construction, illustrated on the spout of a teapot.
    (a) Initialize a straight spine (dashed) from a segmented region (in teal) toward the part's bounding-box center.
    (b) Iterative slicing updates the spine (dashed) and initializes section curves (in orange); red highlights an intersection between section curves.
    (c) Refine the GC to improve the surface fitting.
    (d) In-plane scaling of section curves produces the part's interior geometry (in orange).}
    \label{fig:interior_geometry_construction}
\end{figure}

\noindent
\textbf{Independent interior construction.} \
In this case, a part's interior geometry is independently constructed based on the part's observed geometry (i.e., exterior surface). 
To this end, we first compute a GC that fits the part's exterior surface, and then scale the GC's cross sections to form the part's interior geometry. 

\noindent
{\em Initializing a GC.} \
We initialize the spine curve $\mathbf{c}(u)$ of the GC as a straight line connecting the center of a segmented region of the part to the center of the part's bounding box; see  Figure~\ref{fig:interior_geometry_construction}(a).
 Next, we iteratively slice the part based on the current spine $\mathbf{c}(u)$, compute the center point of each slice, and fit these center points using a B-spline curve to update the spine $\mathbf{c}(u)$ until convergence.
 We initialize each section curve $\mathbf{q}(u, v)$ as a planar curve that fits the boundary of each slice. 
 Figure~\ref{fig:interior_geometry_construction}(b) shows an initial GC constructed using this approach.

\noindent
{\em Refining a GC.} \
The initial GC may not fit the part's exterior surface well and it may have  intersections among neighboring cross sections, since the initial section curves are planar.
To address this limitation, we refine the GC's geometry using an optimization that minimizes
\begin{equation}
\min_{\mathbf{c},\mathbf{q}}\;
\mathcal{E}_{\mathrm{fit}}
+\lambda_s\mathcal{E}_{\mathrm{smooth}}
+\lambda_f\mathcal{E}_{\mathrm{fold}},
\label{eq:gc_optimization}
\end{equation}
where
$\mathcal{E}_{\mathrm{fit}}$ measures agreement with the part's observed geometry, 
$\mathcal{E}_{\mathrm{smooth}}$ penalizes rapid longitudinal changes in the spine curve and the surface, and $\mathcal{E}_{\mathrm{fold}}$ discourages local folding.
The nonnegative weights $\lambda_s$ and $\lambda_f$ balance these terms.
Note that when a straight spine curve is sufficient to capture the part's longitudinal shape during initialization, we keep it straight during optimization by adding a hard constraint.
Figure~\ref{fig:interior_geometry_construction}(c) shows a GC after the refinement.
Please refer to Section~S4 of the supplementary material for details of the optimization.

\noindent
{\em Scaling a GC.} \
We obtain the part's interior geometry by scaling each cross section of the GC in its local reference plane.
This is achieved by replacing $\mathbf{q}(u,v)=(x,y,z)^{\mathsf T}$ with $(\alpha x,\alpha y,z)^{\mathsf T}$, where $0<\alpha<1$.
In our experiments, $\alpha$ is typically set a value in the range [0.75, 0.85], defining the shell thickness of the part.
Figure~\ref{fig:interior_geometry_construction}(d) shows interior geometry constructed for a spout.

\begin{figure}[tbp]
    \centering
    \includegraphics[width=\linewidth]{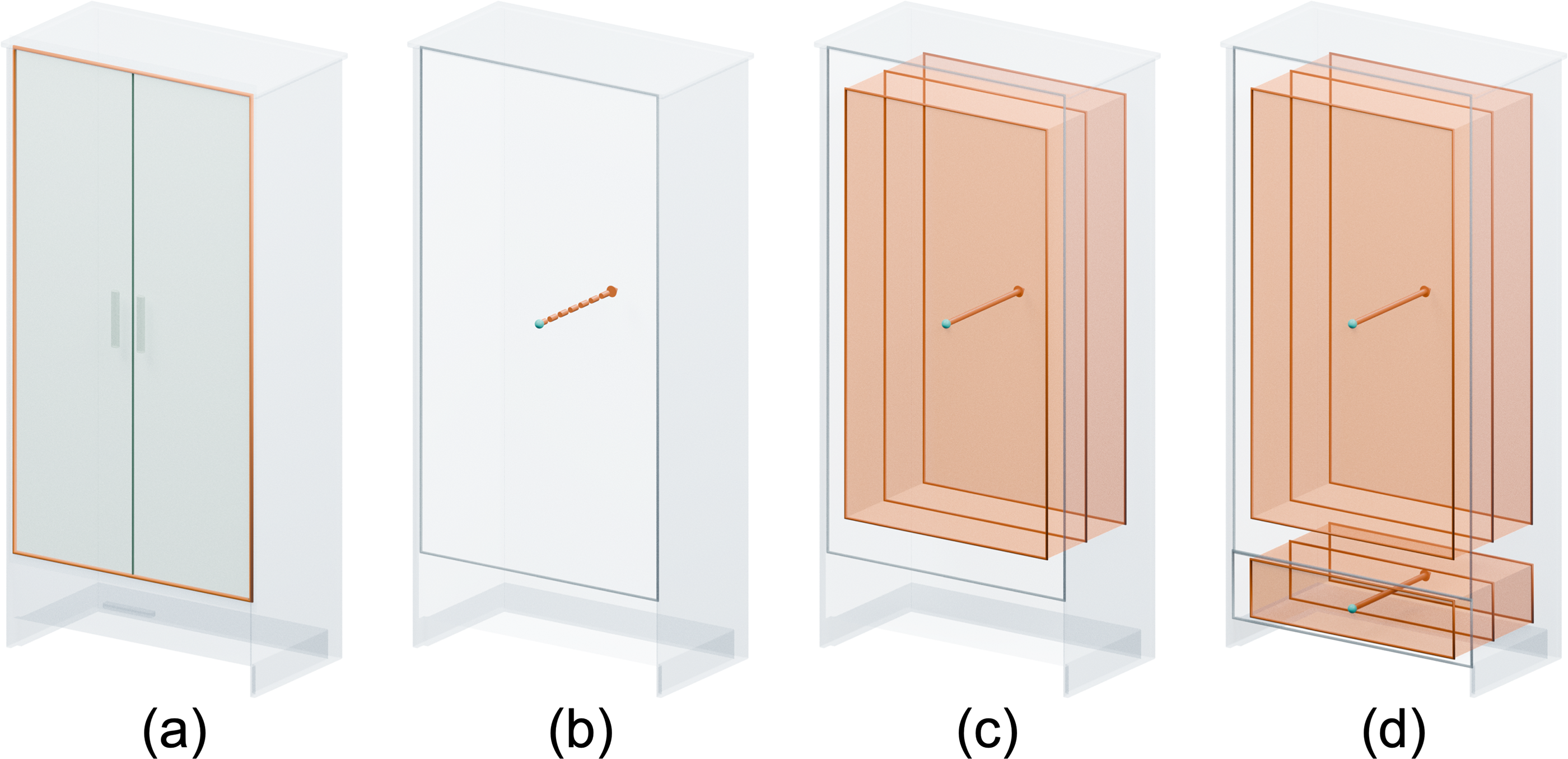}
    \caption{Dependent interior construction, illustrated on the body of a wardrobe.
    	(a) A group of source parts define a combined entrance outline. 
    	(b) The spine is initialized at the outline centroid along the inward plane normal. 
    	(c) Sweeping the inward-scaled outline produces a constant-section GC with target-bounded depth and rear-wall clearance.
    (d) Complete interior geometry constructed for the body part.
    	}
    \label{fig:associated_part_construction}
\end{figure}

\noindent
\textbf{Dependent interior construction.} \
In this case, our goal is to construct the interior geometry of a part that is the target node of $\mathit{define}$ edges in the part graph, where the interior geometry is defined by the target part as well as the associated source parts; see the body part of a wardrobe in Figure~\ref{fig:part_graph_construction}(b) for an example.  
 
We construct a GC to specify the space to be carved from this target part by each group of source parts.
 To this end, for a group $g_l$, we project its observed boundaries onto a common reference plane to obtain a combined outline, as shown in Figure~\ref{fig:associated_part_construction}(a).
 Let $\mathbf{q}_g(v)$ denote this outline in local coordinates relative to its area centroid.
 We initialize a straight spine at this centroid, directed into the target along the plane normal, as illustrated in Figure~\ref{fig:associated_part_construction}(b). 
 We update the spine as described above for independent interior construction, using the corresponding target region while keeping the entrance position fixed.
 Next, we scale the outline inward and sweep the resulting constant section along the spine, with $\mathbf{q}(u,v)=\alpha_g\mathbf{q}_g(v)$ for all $u\in[0,1]$ and $0<\alpha_g<1$.
 The target part's geometry limits the GC's inward extent, leaving clearance for the rear wall, as shown in Figure~\ref{fig:associated_part_construction}(c).
We repeat this process for each group of source parts associated with the target part. 
 Figure~\ref{fig:associated_part_construction}(d) shows the complete interior geometry of a target part constructed using this approach.

\begin{figure}[tbp]
	\centering
	\includegraphics[width=\linewidth]{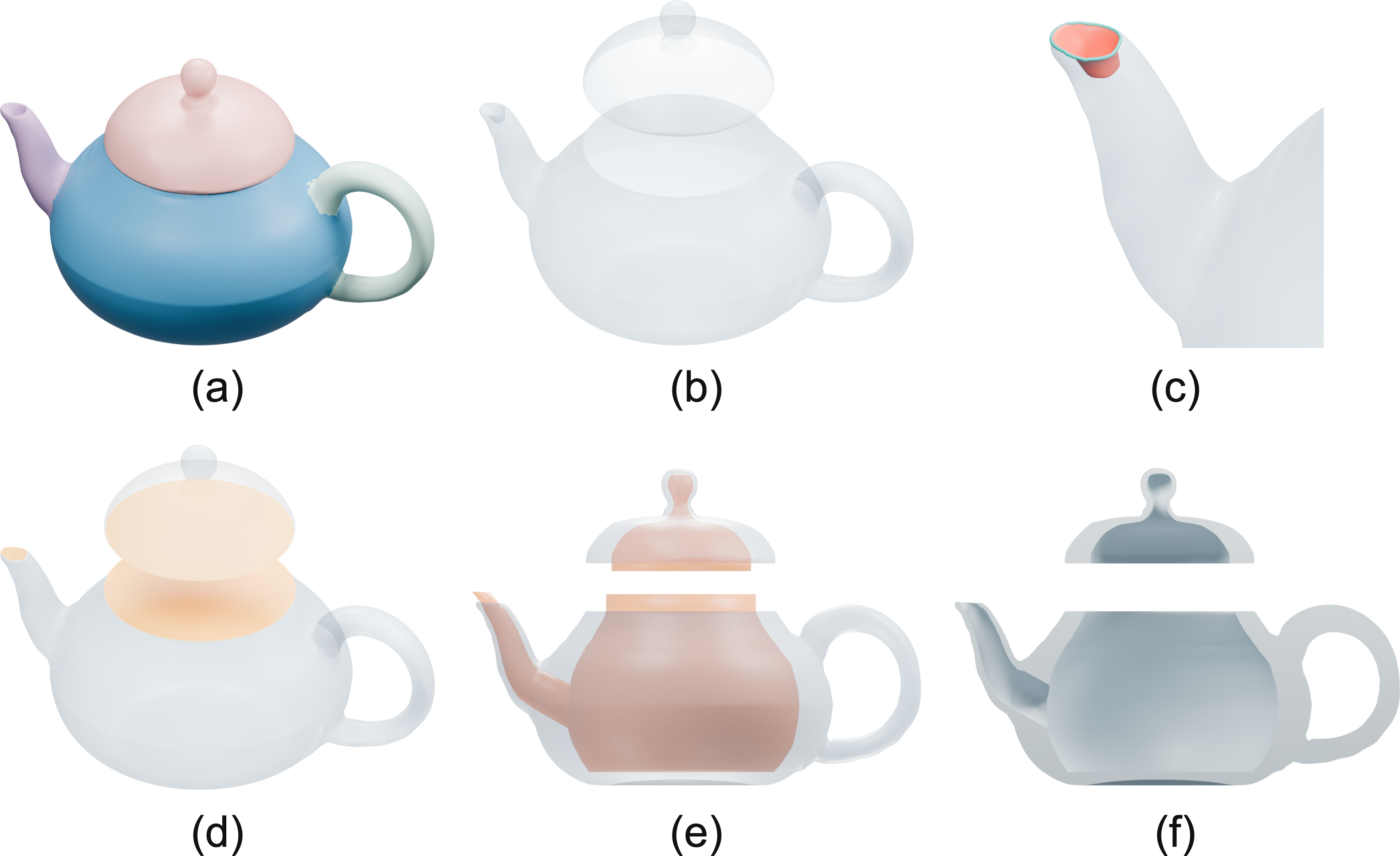}
	\caption{Constructing the output shape for an AI-generated teapot.
		(a) VLM-guided shape segmentation.
		(b) Two solids identified (i.e., body+spout+handle, lid).
		(c) Residual interior surfaces near the spout opening (in red).
		(d) Two solids with boundary closed (in yellow).
		(e) Extended GC volumes (in orange), where the GCs in the body and spout are united.
		(f) Output shape after GC volume subtraction, shown in a cutaway view.
		\vspace*{-2mm}
		}
	\label{fig:surface_preparation_assembly}
\end{figure}

\subsection{Output Shape Construction}
\label{sec:assembly}
The GCs constructed in Section~\ref{sec:interior_geometry_construction} specify the interior spaces within individual parts.
To construct the output shape, we prepare a collection of closed solids $\{O_r\}$ from the parts' exterior surfaces.
For each solid $O_r$ requiring subtraction, we construct a subtraction volume $V_r$ from the corresponding GC(s). 
The resulting shape $S_r$ is
\begin{equation}
S_r=O_r\setminus V_r.
\label{eq:assembly_boolean}
\end{equation}

To prepare the set of solids $\{O_r\}$, we start from the segmented parts in Section~\ref{subsec:shape_segmentation} and merge parts that are connected by $\mathit{union}$ and $\mathit{join}$ edges in the part graph. 
Parts connected by $\mathit{separate}$ edges will form different solids. 
 Figure~\ref{fig:surface_preparation_assembly}(b) shows two solids (i.e., body+handle+spout, lid) are constructed for a teapot example.
In each solid, we remove the AI-generated residual interior surface (if any), usually near one end of a GC.
Figure~\ref{fig:surface_preparation_assembly}(c\&d) shows the residual interior surface near the spout opening is removed.
We then close the remaining open boundaries for each solid, including the boundaries that are caused by segmenting two parts connected by a $\mathit{separate}$ edge in the part graph,  as illustrated by the teapot body and lid in Figure~\ref{fig:surface_preparation_assembly}(d).

To construct the subtraction volume $V_r$ for each solid $O_r$, we process the relevant GCs according to the intended openings and edge attributes, using the volumes enclosed after capping their ends.
At intended openings, we extend the GC beyond the outer surface so that subtraction creates an opening, as illustrated in Figure~\ref{fig:surface_preparation_assembly}(e).
Where an end wall is required, such as at the bottom of the teapot body, we stop the GC short of the outer surface.
For $\mathit{define}$ edges, the GC constructed for each group of source parts forms a subtraction volume $V_r$ for the target part.
For a $\mathit{join}$ edge, we extend the GC of the source part toward the GC of the target part until their enclosed volumes overlap.
We unite these overlapping volumes when constructing $V_r$, aim to create a connected interior space after subtraction; see the connected interior space in the spout and body in Figure~\ref{fig:surface_preparation_assembly}(e).
Subtracting the corresponding $V_r$ from each prepared solid $O_r$ yields the resulting shape shown in Figure~\ref{fig:surface_preparation_assembly}(f).

\begin{table}[t]
    \centering
    \caption{Statistics and runtime breakdown for all results in Figures~\ref{fig:result_wateringPot_variaty} and~\ref{fig:result_shape}. Parentheses in Fig. indicate rows counted from top to bottom. $|V|+|E|$ gives part graph node and edge counts; $|V_{\mathrm{in}}|$ counts nodes with $h_i=1$. $T_{\text{gene}}$, $T_{\text{unde}}$, and $T_{\text{proc}}$ denote the time spent on shape generation, shape understanding, and shape processing, respectively; $T_{\text{total}}$ is their sum. All times are in minutes.
    }
    \label{tab:result_statistics}
    \footnotesize
    \setlength{\tabcolsep}{1.5pt}
    \resizebox{\columnwidth}{!}{%
    \begin{tabular}{@{}llrrrrrr@{}}
        \hline
        Fig. & Object & $|V|+|E|$ & $|V_{\mathrm{in}}|$ & $T_{\text{gene}}$ & $T_{\text{unde}}$ & $T_{\text{proc}}$ & $T_{\text{total}}$ \\
        \hline
        \ref{fig:result_wateringPot_variaty}(1) & Watering can 1 & 3 + 2 & 3 & 1.5 & 2.8 & 0.6 & 4.9 \\
        \ref{fig:result_wateringPot_variaty}(2) & Watering can 2 & 5 + 6 & 5 & 2.3 & 3.3 & 1.0 & 6.6 \\
        \ref{fig:result_wateringPot_variaty}(3) & Watering can 3 & 3 + 2 & 3 & 2.5 & 3.0 & 0.7 & 6.2 \\
        \ref{fig:result_wateringPot_variaty}(4) & Watering can 4 & 4 + 4 & 4 & 2.2 & 3.1 & 0.8 & 6.1 \\
        \ref{fig:result_wateringPot_variaty}(5) & Watering can 5 & 6 + 7 & 5 & 2.5 & 3.5 & 1.1 & 7.1 \\
        \hline
        \ref{fig:result_shape}(1) & Vase & 1 + 0 & 1 & 1.2 & 0.5 & 0.2 & 1.9 \\
        \ref{fig:result_shape}(3) & Ring vase & 3 + 2 & 3 & 1.7 & 1.4 & 0.6 & 3.7 \\
        \ref{fig:result_shape}(4) & Teapot & 4 + 3 & 3 & 2.9 & 3.0 & 0.7 & 6.6 \\
        \ref{fig:result_shape}(5) & Wardrobe & 4 + 6 & 2 & 1.8 & 2.9 & 0.4 & 5.1 \\
        \ref{fig:result_shape}(6) & Cabinet & 6 + 7 & 2 & 2.5 & 3.4 & 0.4 & 6.3 \\
        \ref{fig:result_shape}(7) & Stove & 4 + 3 & 3 & 2.4 & 3.0 & 0.6 & 6.0 \\
        \hline
    \end{tabular}%
    }
\end{table}

\begin{figure}[!t]
	\centering
	\includegraphics[width=\linewidth]{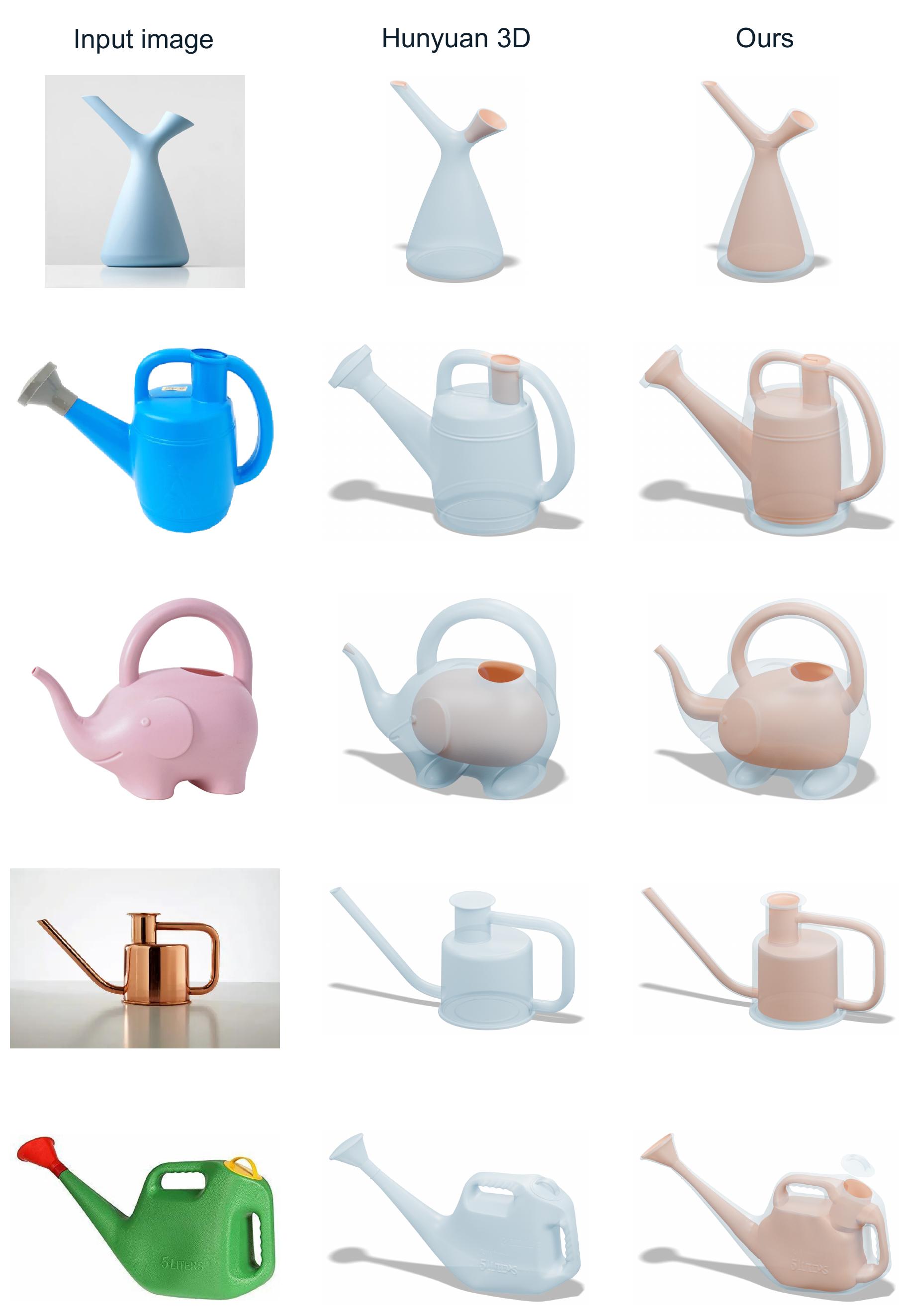} 
	\caption{Our approach is able to functionalize shapes in the same object category (i.e., watering can) yet with different structures and forms, generated using an  image-to-3D AI model~\cite{2026-Hunyuan3D}.
	}
	\label{fig:result_wateringPot_variaty}
\end{figure}

\begin{figure}[!t]
	\centering
	\includegraphics[width=\linewidth]{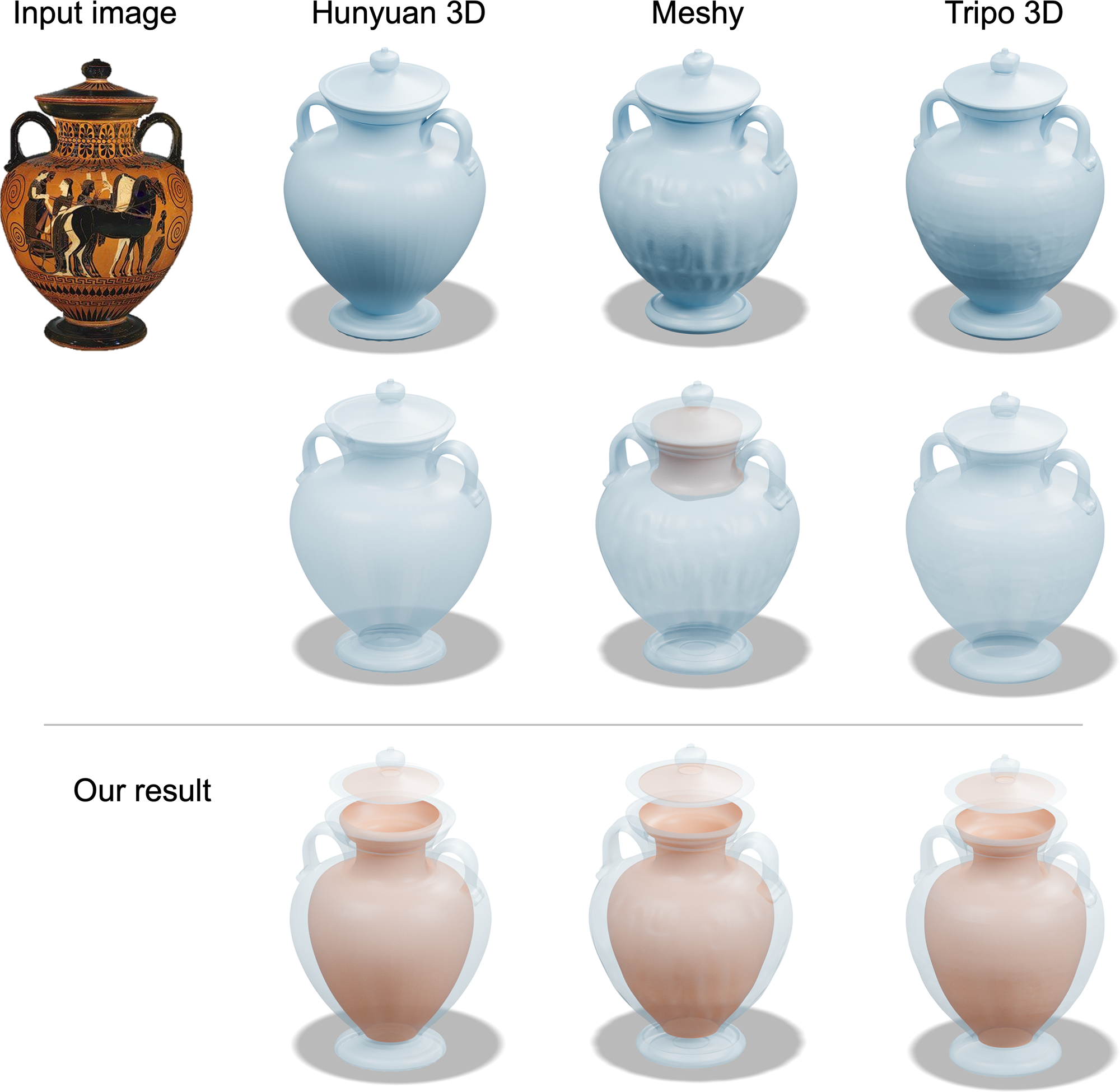}
	\caption{Our approach is able to functionalize shapes generated by different image-to-3D AI models.
	}
	\label{fig:result_AI_image}
\end{figure}

\section{Results}
\label{sec:result}

\paragraph*{Implementation.}
We implemented our method in Python on a machine with a 16-core Intel CPU at 3.0GHz, 32GB RAM, and an RTX 4090 GPU with 24GB VRAM. We use NumPy~\cite{Harris-2020-NumPy} and SciPy~\cite{Virtanen-2020-SciPy} for numerical computation, Manifold~\cite{Manifold-Library} for mesh Boolean operations, and CGAL~\cite{CGAL-Project} for geometry processing and mesh validation.
Given an image or a text prompt, we generate the input shape using Hunyuan3D~\cite{2026-Hunyuan3D}, Meshy~\cite{2026-Meshy}, or Tripo 3D~\cite{2026-Tripo3D}, and obtain an initial segmentation using P3-SAM~\cite{Ma-2026-P3-SAM}.
We access Qwen3.7-Plus through the Qwen API~\cite{Qwen-API} for semantic interpretation and part graph construction, with the temperature set to zero.
Reasoning is disabled by default and enabled for difficult cases.
We solve the optimization in Section~\ref{sec:interior_geometry_construction} using SciPy's nonlinear least-squares solver.

\paragraph*{Results.}
We show that our approach allows functionalizing 
shapes of different object categories, 
shapes in the same category yet with different structures, and
shapes generated by different AI models.

\noindent
\emph{Shapes of different object categories.}
Figure~\ref{fig:teaser} presents shapes of a variety of object categories functionalized using our approach, including 
vases, teapots, desks, a wardrobe, a cabinet, a treasure chest, a trash bin, a violoncello case, and a pipe.
These objects have different interior spaces, such as cavities accessible through openings, connected passages for pouring, and storage compartments behind doors and drawers.
Our approach constructs the interior space successfully for each AI-generated shape to make it functional.

\noindent
\emph{Shapes in the same object category.}
Everyday objects in the same category can have different structures and forms. 
Figure~\ref{fig:result_wateringPot_variaty} shows watering cans that vary in the number and arrangement of their spouts and hollow handles, as well as how these parts connect to the body or to one another.
Our interior-aware part graph captures information necessary for interior geometry construction.
Guided by the graph, our approach is able to successfully construct interior geometry for each watering can.

\begin{figure}[!t]
	\centering
	\includegraphics[width=\linewidth]{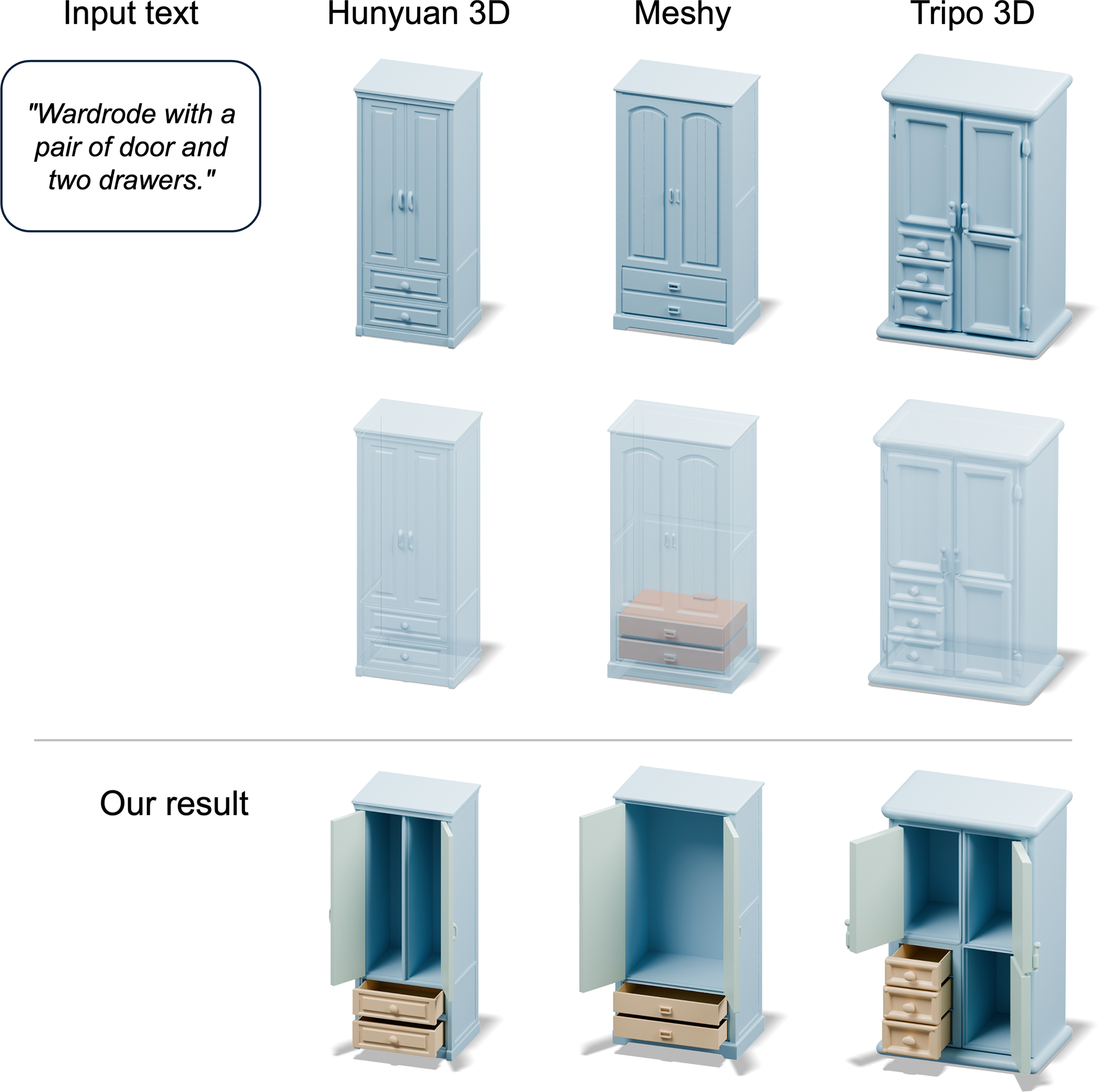}
	\caption{Our approach is able to functionalize shapes generated by different text-to-3D AI models.
	}
	\label{fig:result_AI_text}
\end{figure}

\noindent
\emph{Shapes generated by different AI models.}
Figure~\ref{fig:result_AI_image} shows vase shapes generated from an image prompt using three different AI models, i.e., Hunyuan3D~\cite{2026-Hunyuan3D}, Meshy~\cite{2026-Meshy}, or Tripo 3D~\cite{2026-Tripo3D}.
Although the three AI models produce visually appealing vase shapes, the interior geometry is either missing or incomplete. 
Our approach constructs interior geometry successfully for all the three AI-generated vase shapes. 
Figure~\ref{fig:result_AI_text} shows wardrobe shapes  generated from a text prompt using the three AI models.
Although the three AI-generated wardrobe shapes have different structures and forms, our approach constructs interior geometry successfully for all of them. 
 
\begin{figure*}[!t]
	\centering
	\includegraphics[width=\textwidth]{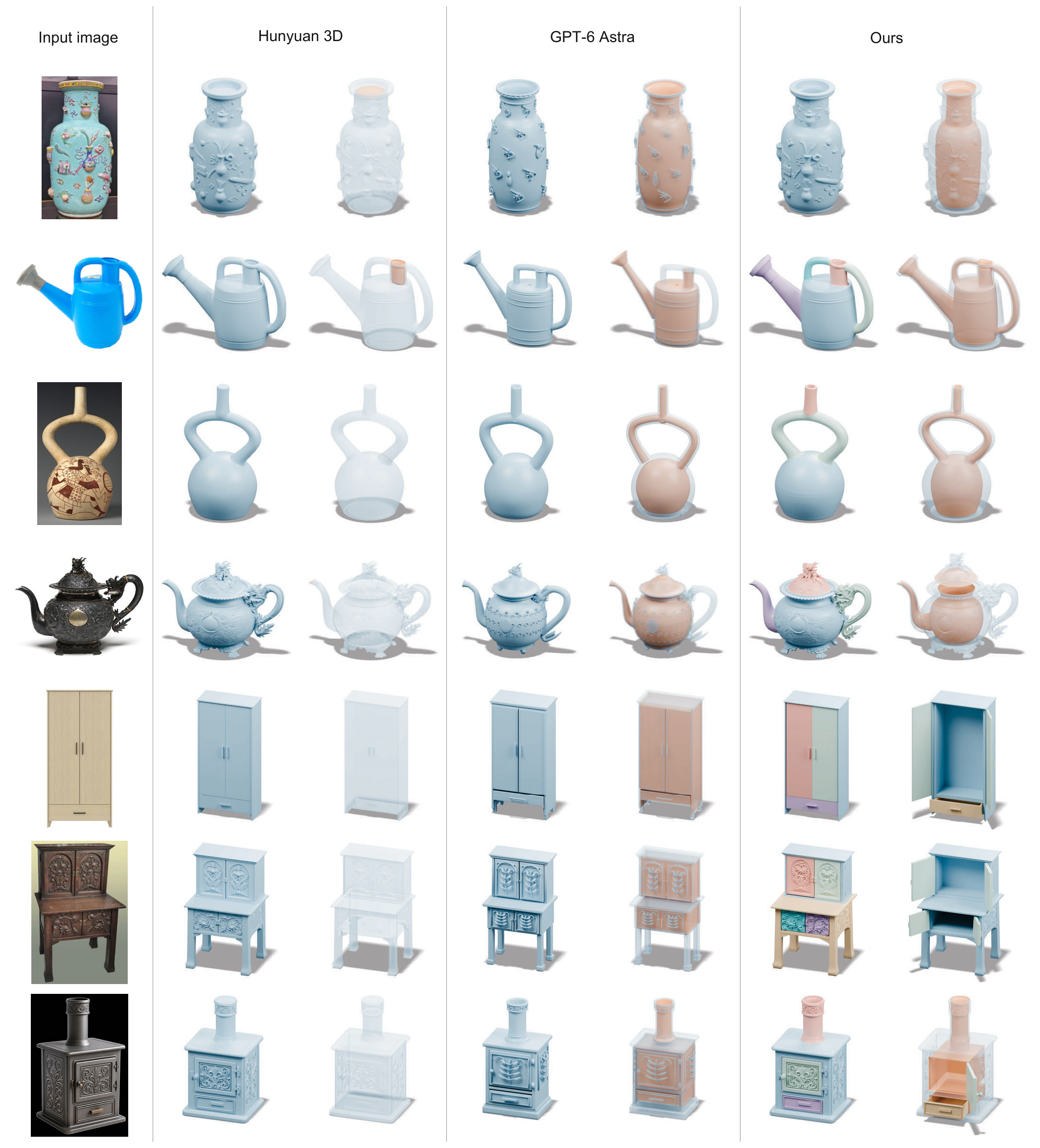}
	\caption{Comparison with direct 3D generation using Hunyuan 3D and GPT-6 Astra. For each AI-generated shape, we show its exterior and interior. For our approach, we show the VLM-based shape segmentation and the functionalized shape.
	}
	\label{fig:result_shape}
\end{figure*}

\paragraph*{Statistics.}
Table~\ref{tab:result_statistics} reports graph node and edge counts, the number of nodes requiring interior space, and a runtime breakdown for all results in Figures~\ref{fig:result_wateringPot_variaty} and~\ref{fig:result_shape}.
$T_{\text{unde}}$ measures the time spent on shape understanding, including VLM-guided shape segmentation and interior-aware part graph construction, which takes around 1 to 3 minutes.
$T_{\text{proc}}$ measures the time spent on shape processing, including part interior geometry construction and output shape construction,  which generally takes less than 1 minute.

\paragraph*{Comparison with direct AI-based 3D generation.}
Figure~\ref{fig:result_shape} compares our approach with Hunyuan 3D~\cite{2026-Hunyuan3D} and GPT-6 Astra~\cite{2026-GPT6Astra}.
Hunyuan 3D produces detailed exteriors but does not provide functional interiors.
GPT-6 Astra usually produces plausible interiors, but cannot guarantee their correctness.
For example,  the central tube in the vase of Figure~\ref{fig:result_shape} (the third row) is not internally connected to the body.
One limitation of GPT-6 Astra is that its generated shape can only approximate freeform geometry in the prompt rather than exactly reproduce it as Hunyuan 3D does.
In contrast,  the shapes functionalized by our approach have not only functional interiors but also visually appealing exteriors produced by the AI model (i.e., Hunyuan 3D). 
Please refer to the supplementary material for an interactive demo of our results.

\begin{figure}[!t]
	\centering
	\includegraphics[width=\linewidth]{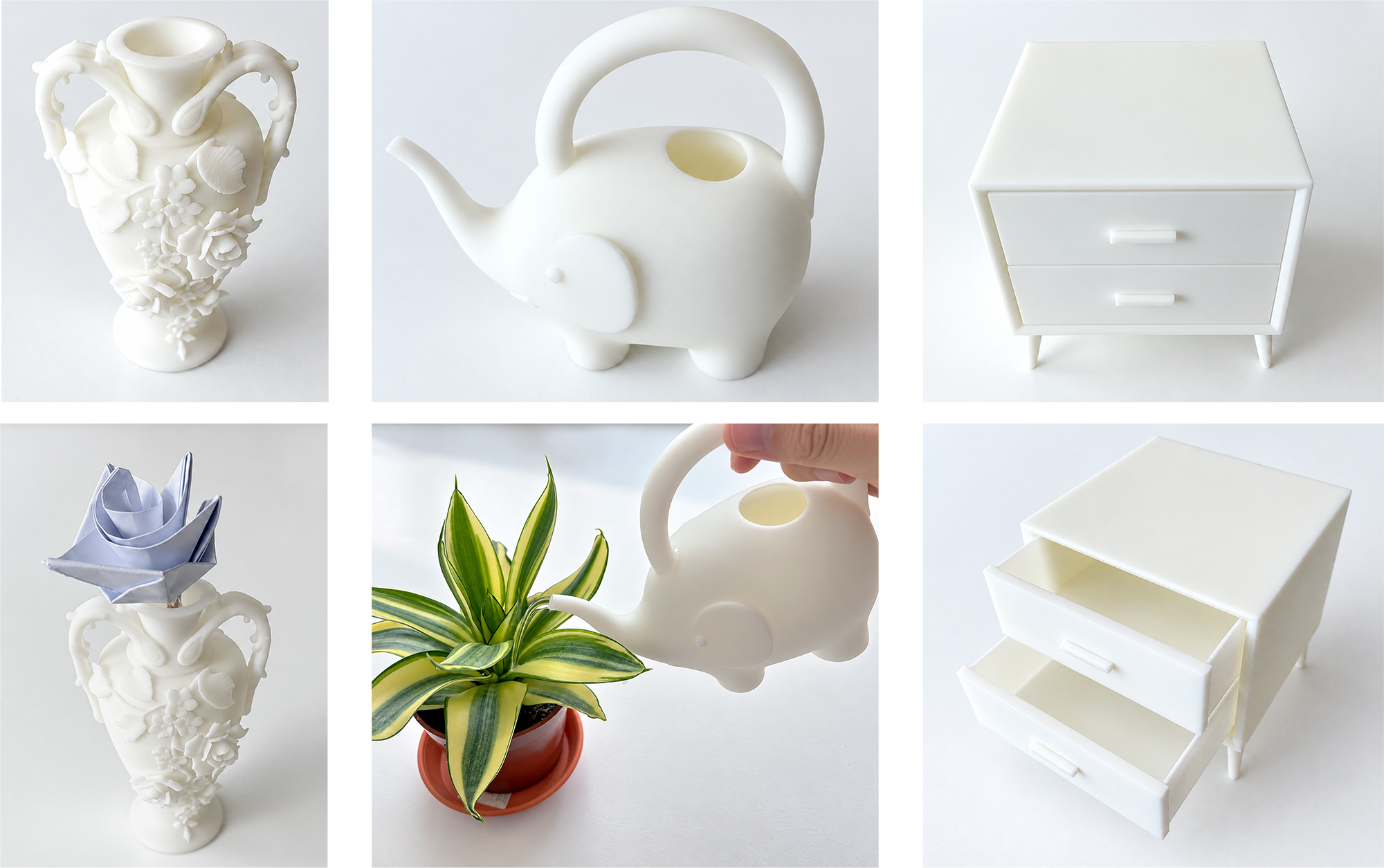}
	\caption{
	3D-printed results for functional validation: from left to right, a decorative vase, a watering can, and a cabinet with pull-out drawers.
	}
	\label{fig:result_fabrication}
\end{figure}

\paragraph*{Fabrication.}
We fabricated three examples for functional validation.
In detail, we 3D printed a decorative vase, a watering can, and a cabinet;  see Figure~\ref{fig:result_fabrication} (top).
Figure~\ref{fig:result_fabrication} (bottom) shows these objects are usable in practice.
The vase provides an open cavity for holding a paper flower, while the watering can's connected body and spout interiors allow water to be poured onto a plant.
The cabinet has separate hollow drawers that fit within its body and can be pulled out to access their storage spaces.

\section{Conclusion}
\label{sec:conclusion}
We presented a computational approach to functionalizing AI-generated 3D shapes through interior geometry construction while preserving their external appearance.
Our approach has two key components: 
building an interior-aware part graph via shape understanding, and 
interior geometry construction guided by the part graph.
The first builds an interior-aware part graph through VLM-guided shape segmentation and graph attribute inference.
The second uses generalized cylinders and geometric optimization to construct part interiors from the observed geometry, and composes them to form the output shape.
We demonstrated our approach on everyday objects with diverse shapes and structures, generated by different AI models.

\noindent
\textbf{Limitations and future work.} 
Our approach has several limitations that open up directions for future work.
First, our geometric construction relies on generalized cylinders and their combinations.
Interiors without a clear sectional structure may require more general geometric representations.
Second, our approach focuses on the interior geometry construction, without explicitly optimizing material strength and mechanical fit.
Integrating physical simulation and fabrication requirements would extend its applicability to practical use.
Lastly, our work focuses on objects whose functionality depends on their interior space.
In the future, we plan to extend  the scope of objects for functionalization, e.g.,  objects whose functionality depends on mechanical joints.
\bibliographystyle{alpha-doi}
\bibliography{Ref_FunctionalShape}

@STRING{TOG_SIG = {ACM Transactions on Graphics (Proc. of SIGGRAPH)}}

@STRING{TOG_SIGA = {ACM Transactions on Graphics (Proc. of SIGGRAPH Asia)}}

@STRING{SIGGRAPH = {Proc. of SIGGRAPH}}

@STRING{SIGGRAPHAsia = {Proc. of SIGGRAPH Asia}}

@STRING{CAD = {Computer-Aided Design}}

@STRING{CHI = {Proc. of ACM CHI}}

@STRING{UIST = {Proc. of ACM UIST}}

@STRING{ICCV = {Proc. of the International Conference on Computer Vision}}

@STRING{CVPR = {Proc. of the IEEE/CVF Conference on Computer Vision and Pattern Recognition}}

@STRING{ECCV = {Proc. of the European Conference on Computer Vision}}

@STRING{AAAI = {Proc. of the AAAI Conference on Artificial Intelligence}}

@STRING{NeurIPS = {Proc. of the Conference on Neural Information Processing Systems}}

@STRING{ICLR = {Proc. of the International Conference on Learning Representations}}

@misc{2026-Hunyuan3D,
	author       =    {{Hunyuan3D}},
	title        =    {{Hunyuan3D}: {AI}-Based {3D} Model Generator},
	url          =    {https://hy3d.tencent.ai/},
	year      =     {2026},
	note     =      {Accessed September 2026},
}

@misc{2026-Meshy,
	author       =    {{Meshy}},
	title        =      {Meshy: {AI}-Based {3D} Model Generator},
	url          =    {https://www.meshy.ai/},
	year      =     {2026},
	note     =      {Accessed September 2026},
}

@misc{2026-Tripo3D,
	author       =    {{Tripo 3D}},
	title        =    {{Tripo 3D}: {AI}-Based {3D} Model Generator},
	url          =    {https://studio.tripo3d.ai/},
	year      =     {2026},
	note     =      {Accessed September 2026},
}

@misc{2026-GPT6Astra,
	author     =    {{OpenAI}},
	title         =    {{GPT-6} {Astra}: A New Generation of Intelligence},
	url           =    {https://openai.com/index/gpt-6-astra/},
	year        =    {2026},
}

@misc{2026-Claude,
	author     =    {{Anthropic}},
	title         =    {Claude},
	url           =    {https://www.anthropic.com/claude},
	year      =     {2026},
}

@book{Ballard-1982-ComputerVision,
	author       =    {Dana H. Ballard and Christopher M. Brown},
	title        =    {Computer Vision},
	publisher    =    {Prentice-Hall},
	year         =    {1982},
}

@article{Zhou-2016-MeshArrangements,
	author       =    {Qingnan Zhou and Eitan Grinspun and Denis Zorin and Alec Jacobson},
	title        =    {Mesh Arrangements for Solid Geometry},
	journal      =    TOG_SIG,
	volume       =    {35},
	number       =    {4},
	pages        =    {39:1--39:15},
	year         =    {2016},
}

@inproceedings{Ma-2026-P3-SAM,
	author     =     {Ma, Changfeng and Li, Yang and Yan, Xinhao and Xu, Jiachen and Yang, Yunhan and Wang, Chunshi and Zhao, Zibo and Guo, Yanwen and Chen, Zhuo and Guo, Chunchao},
	title         =     {{P3-SAM}: Native {3D} Part Segmentation},
	booktitle = ECCV,
	year        =     {2026},
}

@article{Harris-2020-NumPy,
	author = {Harris, Charles R. and Millman, K. Jarrod and van der Walt, St\'efan J. and others},
	title = {Array programming with {NumPy}},
	journal = {Nature},
	volume = {585},
	pages = {357--362},
	year = {2020},
}

@article{Virtanen-2020-SciPy,
	author = {Virtanen, Pauli and Gommers, Ralf and Oliphant, Travis E. and others},
	title = {{SciPy} 1.0: Fundamental Algorithms for Scientific Computing in {Python}},
	journal = {Nature Methods},
	volume = {17},
	pages = {261--272},
	year = {2020},
}

@misc{Manifold-Library,
	author = {{Manifold Contributors}},
	title = {{Manifold}: Geometry Library for Topological Robustness},
	url = {https://github.com/elalish/manifold},
	year = {2026},
}

@book{CGAL-Project,
	title        = "{CGAL} User and Reference Manual", 
	author      = "{The CGAL Project}", 
	publisher     = "{CGAL Editorial Board}", 
	edition     = "{6.2.1}", 
	year         = 2026, 
	url =    "https://doc.cgal.org/6.2.1/Manual/packages.html"
}

@misc{Qwen-API,
	author = {{Alibaba Cloud}},
	title = {{Alibaba Cloud Model Studio}: Text Generation {API} Reference},
	year = {2026},
	url = {https://www.alibabacloud.com/help/en/model-studio/qwen-api-reference/}
}

@inproceedings{Poole-2023-DreamFusion,
	author       =    {Ben Poole and Ajay Jain and Jonathan T. Barron and Ben Mildenhall},
	title            =    {{DreamFusion}: Text-to-{3D} using {2D} Diffusion},
	booktitle    =    ICLR,
	year          =    {2023},
}

@inproceedings{Wang-2023-ProlificDreamer,
	author       =    {Zhengyi Wang and Cheng Lu and Yikai Wang and Fan Bao and Chongxuan Li and Hang Su and Jun Zhu},
	title            =    {{ProlificDreamer}: High-Fidelity and Diverse Text-to-3D Generation with Variational Score Distillation},
	booktitle    =    NeurIPS,
	volume       =    {36},
	pages        =     {8406--8441},
	year           =    {2023},
}

@inproceedings{Tang-2024-DreamGaussian,
	author       =    {Jiaxiang Tang and Jiawei Ren and Hang Zhou and Ziwei Liu and Gang Zeng},
	title            =    {{DreamGaussian}: Generative Gaussian Splatting for Efficient 3D Content Creation},
	booktitle    =    ICLR,
    year           =    {2024},
}

@inproceedings{Lin-2023-Magic3D,
	author       =    {Chen-Hsuan Lin and Jun Gao and Luming Tang and Towaki Takikawa and Xiaohui Zeng and Xun Huang and Karsten Kreis and Sanja Fidler and Ming-Yu Liu and Tsung-Yi Lin},
	title            =    {{Magic3D}: High-Resolution Text-to-{3D} Content Creation},
	booktitle    =    CVPR,
	pages        =    {300--309},
	year          =    {2023},
}

@misc{Xu-2024-InstantMesh,
	author       =    {Jiale Xu and Weihao Cheng and Yiming Gao and Xintao Wang and Shenghua Gao and Ying Shan},
	title        =    {{InstantMesh: Efficient 3D Mesh Generation from a Single Image with Sparse-view Large Reconstruction Models}},
	year         =    {2024},
	eprint       =    {2404.07191},
	archivePrefix=   {arXiv},
	primaryClass =   {cs.CV},
}

@inproceedings{Liu-2023-Zero123,
	author       =    {Ruoshi Liu and Rundi Wu and Basile {Van Hoorick} and Pavel Tokmakov and Sergey Zakharov and Carl Vondrick},
	title            =    {{Zero-1-to-3}: Zero-shot One Image to {3D} Object},
	booktitle    =    ICCV,
	pages        =    {9298--9309},
	year          =    {2023},
}

@inproceedings{Shi-2023-MVDream,
	author       =    {Yichun Shi and Peng Wang and Jianglong Ye and Long Mai and Kejie Li and Xiao Yang},
	title        =    {{MVDream: Multi-view Diffusion for 3D Generation}},
	booktitle    =    ICLR,
	year         =    {2024},
}

@inproceedings{Long-2023-Wonder3D,
	author       =    {Xiaoxiao Long and Yuan-Chen Guo and Cheng Lin and Yuan Liu and Zhiyang Dou and Lingjie Liu and Yuexin Ma and Song-Hai Zhang and Marc Habermann and Christian Theobalt and Wenping Wang},
	title        =    {{Wonder3D: Single Image to 3D using Cross-Domain Diffusion}},
	booktitle    =    CVPR,
	pages        =    {9970--9980},
	year         =    {2024},
}

@misc{Xu-2026-Meshy,
	author    =    {Jiale Xu and Rendong Liang and Yuhao Long and Siyuan Shen and Zangyueyang Xian and Xiaofei Wu and Zeyi Xu and Yuanming Hu},
	title        =    {{Meshy T2}: Fast Native Mesh Generation with Flow Matching}, 
	year       =    {2026},
	eprint     =    {2607.28675},
	archivePrefix = {arXiv},
	primaryClass = {cs.GR},
}

@misc{Zhao-2025-Hunyuan3D,
	author       =    {Zibo Zhao and Zeqiang Lai and Qingxiang Lin and Yunfei Zhao and Haolin Liu and Shuhui Yang and Yifei Feng and Mingxin Yang and Sheng Zhang and Xianghui Yang and Huiwen Shi and Sicong Liu and Junta Wu and Yihang Lian and Fan Yang and Ruining Tang and Zebin He and Xinzhou Wang and Jian Liu and Xuhui Zuo and Zhuo Chen and Biwen Lei and Haohan Weng and Jing Xu and Yiling Zhu and Xinhai Liu and Lixin Xu and Changrong Hu and Tianyu Huang and Lifu Wang and Jihong Zhang and Meng Chen and Liang Dong and Yiwen Jia and Yulin Cai and Jiaao Yu and Yixuan Tang and Hao Zhang and Zheng Ye and Peng He and Runzhou Wu and Chao Zhang and Yonghao Tan and Jie Xiao and Yangyu Tao and Jianchen Zhu and Jinbao Xue and Kai Liu and Chongqing Zhao and Xinming Wu and Zhichao Hu and Lei Qin and Jianbing Peng and Zhan Li and Minghui Chen and Xipeng Zhang and Lin Niu and Paige Wang and Yingkai Wang and Haozhao Kuang and Zhongyi Fan and Xu Zheng and Weihao Zhuang and YingPing He and Tian Liu and Yong Yang and Di Wang and Yuhong Liu and Jie Jiang and Jingwei Huang and Chunchao Guo},
	title        =    {{Hunyuan3D 2.0}: Scaling Diffusion Models for High Resolution Textured {3D} Assets Generation},
	year         =    {2025},
	eprint       =    {2501.12202},
	archivePrefix=   {arXiv},
	primaryClass =   {cs.CV},
}

@misc{Tochilkin-2024-TripoSR,
	author       =    {Dmitry Tochilkin and David Pankratz and Zexiang Liu and Zixuan Huang and Adam Letts and Yangguang Li and Ding Liang and Christian Laforte and Varun Jampani and Yan-Pei Cao},
	title        =    {{TripoSR}: Fast {3D} Object Reconstruction from a Single Image},
	year         =    {2024},
	eprint       =    {2403.02151},
	archivePrefix=   {arXiv},
	primaryClass =   {cs.CV},
}

@inproceedings{Xiang-2024-TRELLIS,
	author       =    {Jianfeng Xiang and Zelong Lv and Sicheng Xu and Yu Deng and Ruicheng Wang and Bowen Zhang and Dong Chen and Xin Tong and Jiaolong Yang},
	title        =    {Structured {3D} Latents for Scalable and Versatile {3D} Generation},
	booktitle    =    CVPR,
	pages        =    {21469--21480},
	year         =    {2025},
}

@article{Zhang-2024-CLAY,
	author       =    {Longwen Zhang and Ziyu Wang and Qixuan Zhang and Qiwei Qiu and Anqi Pang and Haoran Jiang and Wei Yang and Lan Xu and Jingyi Yu},
	title        =    {{CLAY}: A Controllable Large-scale Generative Model for Creating High-quality {3D} Assets},
	journal      =    TOG_SIG,
	volume       =    {43},
	number       =    {4},
	pages        =    {120:1--120:20},
	year         =    {2024},
}

@inproceedings{Wu-2024-Direct3D,
	author       =    {Shuang Wu and Youtian Lin and Feihu Zhang and Yifei Zeng and Jingxi Xu and Philip Torr and Xun Cao and Yao Yao},
	title        =    {{Direct3D}: Scalable Image-to-{3D} Generation via {3D} Latent Diffusion Transformer},
	booktitle    =    NeurIPS,
	volume       =    {37},
	pages        =    {121859--121881},
	year         =    {2024},
}

@inproceedings{Hong-2023-LRM,
	author       =    {Yicong Hong and Kai Zhang and Jiuxiang Gu and Sai Bi and Yang Zhou and Difan Liu and Feng Liu and Kalyan Sunkavalli and Trung Bui and Hao Tan},
	title        =    {{LRM: Large Reconstruction Model for Single Image to 3D}},
	booktitle    =    ICLR,
	year         =    {2024},
}

@inproceedings{Zhao-2023-Michelangelo,
	author       =    {Zibo Zhao and Wen Liu and Xin Chen and Xianfang Zeng and Rui Wang and Pei Cheng and Bin Fu and Tao Chen and Gang Yu and Shenghua Gao},
	title        =    {Michelangelo: Conditional {3D} Shape Generation based on Shape-Image-Text Aligned Latent Representation},
	booktitle    =    NeurIPS,
	volume       =    {36},
	pages        =    {73969--73982},
	year         =    {2023},
}

@inproceedings{Liu-2026-LandCraft,
	author       =     {Zhihao Liu and Fang Liu and Weihao Xuan and Naoto Yokoya},
	title            =     {{LandCraft}: Designing the Structured 3D Landscapes via Text Guidance},
	booktitle   =     AAAI,
	volume     =     {40},
	pages       =      {7467--7475},
	year          =      {2026},
}

@inproceedings{Sun-2025-3D-GPT,
	author        =    {Chunyi Sun and Junlin Han and Weijian Deng and Xinlong Wang and Zishan Qin and Stephen Gould},
	title             =    {{3D-GPT}: Procedural {3D} Modeling with Large Language Models},
	booktitle    =    {2025 International Conference on 3D Vision (3DV)},
	pages         =    {1253--1263},
	year            =    {2025},
}

@inproceedings{Zhou-2025-SceneX,
	author         =      {Mengqi Zhou and Yuxi Wang and Jun Hou and Shougao Zhang and Yiwei Li and Chuanchen Luo and Junran Peng and Zhaoxiang Zhang},
	title              =     {{SceneX}: Procedural Controllable Large-Scale Scene Generation},
	booktitle     =     AAAI,
	volume       =     {39},
	pages         =    {10806--10814},
	year           =     {2025},
}

@inproceedings{Liu-2026-MultiAgentCAD,
	author       =     {Liu, Yang and Ren, Daxuan and Ding, Yijie and Zheng, Jianmin and Deng, Fang},
	title            =     {Multi-Agent {CAD} Code Generation},
	booktitle   =     SIGGRAPH,
	pages       =      {20:1--20:13},
	year          =      {2026},
}

@inproceedings{Zhang-2025-ShapeCraft,
	author        =    {Shuyuan Zhang and ChenHan Jiang and Zuoou Li and Jiankang Deng},
	title             =    {{ShapeCraft}: {LLM} Agents for Structured, Textured and Interactive {3D} Modeling},
	booktitle    =    NeurIPS,
	volume       =    {38},
	pages         =    {65116--65144},
	year           =    {2025},
}

@inproceedings{Chen-2025-CADCrafter,
	author       =    {Cheng Chen and Jiacheng Wei and Tianrun Chen and Chi Zhang and Xiaofeng Yang and Shangzhan Zhang and Bingchen Yang and Chuan-Sheng Foo and Guosheng Lin and Qixing Huang and Fayao Liu},
	booktitle  =     CVPR,
	title          =    {{CADCrafter}: Generating Computer-Aided Design Models from Unconstrained Images},
	pages      =     {11073--11082},
	year         =    {2025},
}

@article{Li-2025-LLM4CAD,
	author    =     {Li, Xingang and Sun, Yuewan and Sha, Zhenghui},
	title         =     {{LLM4CAD}: Multimodal Large Language Models for Three-Dimensional Computer-Aided Design Generation},
	journal    =     {ASME Journal of Computing and Information Science in Engineering},
	volume   =     {25},
	number   =     {2},
	pages      =     {021005:1--021005:14},
	year        =     {2025},
}

@inproceedings{Huang-2024-BlenderAlchemy,
	author       =    {Ian Huang and Guandao Yang and Leonidas Guibas},
	title             =    {{BlenderAlchemy}: Editing {3D} Graphics with Vision-Language Models},
	booktitle    =    ECCV,
	series        =    {Lecture Notes in Computer Science},
	volume      =    {15147},
	pages       =    {297--314},
	year           =    {2024},
}

@inproceedings{Hu-2024-SceneCraft,
	author       =       {Ziniu Hu and Ahmet Iscen and Aashi Jain and Thomas Kipf and Yisong Yue and David A. Ross and Cordelia Schmid and Alireza Fathi},
	title            =      {{SceneCraft}: An {LLM} Agent for Synthesizing {3D} Scenes as {Blender} Code},
	booktitle   =      {Proc. of the 41st International Conference on Machine Learning},
	series      =       {Proceedings of Machine Learning Research},
	volume      =       {235},
	pages       =       {19252--19282},
	year          =       {2024},
}

@inproceedings{Mallis-2025-CADAssistant,
	author       =    {Dimitrios Mallis and Ahmet Serda Karadeniz and Sebastian Cavada and Danila Rukhovich and Niki Foteinopoulou and Kseniya Cherenkova and Anis Kacem and Djamila Aouada},
	title        =    {{CAD-Assistant}: Tool-Augmented {VLLM}s as Generic {CAD} Task Solvers},
	booktitle    =    ICCV,
	pages        =    {7284--7294},
	year         =    {2025},
}

@inproceedings{Faruqi-2023-Style2Fab,
	author       =    {Faraz Faruqi and Ahmed Katary and Tarik Hasic and Amira Abdel-Rahman and Nayeemur Rahman and Leandra Tejedor and Mackenzie Leake and Megan Hofmann and Stefanie Mueller},
	title            =    {{Style2Fab}: Functionality-Aware Segmentation for Fabricating Personalized {3D} Models with Generative {AI}},
	booktitle    =    UIST,
	pages        =   {22:1--22:13},
	year         =    {2023},
}

@inproceedings{Faruqi-2024-ShapingRealities,
   author        =     {Faruqi, Faraz and Tian, Yingtao and Phadnis, Vrushank and Jampani, Varun and Mueller, Stefanie},
   title             =     {Shaping Realities: Enhancing {3D} Generative {AI} with Fabrication Constraints},
   booktitle     =     {{GenAICHI}: {CHI} 2024 Workshop on Generative {AI} and {HCI}},
   year            =     {2024},
   pages        =     {1--5},
}

@inproceedings{Dang-2026-DensiCrafter,
	author          =    {Shengqi Dang and Fu Chai and Jiaxin Li and Chao Yuan and Wei Ye and Nan Cao},
	title              =    {{DensiCrafter}: Physically-Constrained Generation and Fabrication of Self-Supporting Hollow Structures},
	booktitle     =    AAAI,
	volume       =      {40},
	pages        =    {193--201},
	year           =    {2026},
}

@inproceedings{Chen-2024-Atlas3D,
	author       =    {Yunuo Chen and Tianyi Xie and Zeshun Zong and Xuan Li and Feng Gao and Yin Yang and Ying Nian Wu and Chenfanfu Jiang},
	title        =    {{Atlas3D}: Physically Constrained Self-Supporting Text-to-{3D} for Simulation and Fabrication},
	booktitle    =    NeurIPS,
	volume       =    {37},
	pages        =    {102501--102530},
	year         =    {2024},
}

@inproceedings{Guo-2024-PhysicallyCompatible,
   author        =     {Guo, Minghao and Wang, Bohan and Ma, Pingchuan and Zhang, Tianyuan and Owens, Crystal Elaine and Gan, Chuang and Tenenbaum, Joshua B. and He, Kaiming and Matusik, Wojciech},
   title            =     {Physically Compatible {3D} Object Modeling from a Single Image},
   booktitle    =     NeurIPS,
   volume       =     {37},
   pages        =    {119260--119282},
   year           =      {2024}
}

@inproceedings{Xu-2025-FunctionsBetter,
	author       =    {Qingshan Xu and Jiao Liu and Melvin Wong and Caishun Chen and Yew Soon Ong},
	title        =    {Looks Great, Functions Better: Physics Compliance Text-to-{3D} Shape Generation},
	booktitle    =    {Proc. of International Joint Conference on Neural Networks (IJCNN)},
	pages        =    {1--10},
	year         =    {2025},
}

@inproceedings{Faruqi-2025-MechStyle,
	author       =    {Faraz Faruqi and Amira Abdel-Rahman and Leandra Tejedor and Martin Nisser and Jiaj Li and Vrushank Phadnis and Varun Jampani and Neil Gershenfeld and Megan Hofmann and Stefanie Mueller},
	title            =    {{MechStyle}: Augmenting Generative {AI} with Mechanical Simulation to Create Stylized and Structurally Viable {3D} Models},
	booktitle    =    {Proc. of the {ACM} Symposium on Computational Fabrication ({SCF})},
	pages        =    {24:1--24:15},
	year           =    {2025},
}

@inproceedings{Zhan-2025-PhysiOpt,
	author       =    {Xiao Zhan and Cl{\'e}ment Jambon and Evan Thompson and Kenney Ng and Mina {Konakovi{\'c} Luki{\'c}}},
	title            =    {{PhysiOpt}: Physics-Driven Shape Optimization for {3D} Generative Models},
	booktitle    =    SIGGRAPHAsia,
	pages        =    {109:1--109:11},
	year           =    {2025},
}

@inproceedings{Botzer-2026-Laserify,
	author        =      {Ziv Botzer and Guy Austern and Yoav Sterman},
	title             =      {Laserify: From Multimodal Input to Fabrication-Ready Laser Cut Plans},
	booktitle     =      {Proc. of Extended Abstracts of the {CHI} Conference on Human Factors in Computing Systems},
	year            =      {2026},
	pages        =      {424:1--424:5},
}

@inproceedings{Pun-2025-BrickStructure,
	author        =     {Pun, Ava and Deng, Kangle and Liu, Ruixuan and Ramanan, Deva and Liu, Changliu and Zhu, Jun-Yan},
	title            =     {Generating Physically Stable and Buildable Brick Structures from Text},
	booktitle    =     ICCV,
	pages        =     {14798--14809},
	year           =      {2025}
}

@inproceedings{Xu-2025-LegoACE,
	author        =    {Hao Xu and Yuqing Zhang and Yiqian Wu and Xinyang Zheng and Yutao Liu and Xiangjun Tang and Yunhan Yang and Ding Liang and Yingtian Liu and Yuanchen Guo and Yanpei Cao and Xiaogang Jin},
	title            =     {{LegoACE}: Autoregressive Construction Engine for Expressive {LEGO}\textsuperscript{\textregistered} Assemblies},
	booktitle   =     SIGGRAPHAsia,
	pages        =     {40:1--40:11},
	year           =     {2025},
}

@article{Ge-2025-LEGOMaker,
	author       =       {Jiahao Ge and Mingjun Zhou and Hanyou Zheng and Hao Xu and Chi-Wing Fu},
	title            =       {{LEGO}\textsuperscript{\textregistered}-Maker: Autoregressive Image-Conditioned {LEGO}\textsuperscript{\textregistered} Model Creation},
	journal       =       TOG_SIGA,
	volume      =       {44},
	number     =       {6},
	pages       =        {201:1--201:15},
	year          =        {2025},
}

@inproceedings{Chen-2025-ArtiLatent,
	author       =    {Honghua Chen and Yushi Lan and Yongwei Chen and Xingang Pan},
	title            =    {{ArtiLatent}: Realistic Articulated {3D} Object Generation via Structured Latents},
	booktitle    =    SIGGRAPHAsia,
	pages        =    {157:1--157:11},
	year           =    {2025},
}

@inproceedings{Le-2025-ArticulateAnything,
	author      =     {Long Le and Jason Xie and William Liang and Hung-Ju Wang and Yue Yang and Yecheng Jason Ma and Kyle Vedder and Arjun Krishna and Dinesh Jayaraman and Eric Eaton},
	title          =     {{Articulate-Anything}: Automatic Modeling of Articulated Objects via a Vision-Language Foundation Model},
	booktitle   =    ICLR,
	year          =     {2025},
}

@inproceedings{Chen-2025-FreeArt3D,
   author       =    {Chuhao Chen and Isabella Liu and Xinyue Wei and Hao Su and Minghua Liu},
   title           =     {{FreeArt3D}: Training-Free Articulated Object Generation using {3D} Diffusion},
   booktitle   =    SIGGRAPHAsia,
   pages       =    {154:1--154:13},
   year          =    {2025},
}

@inproceedings{Lu-2025-DreamArt,
	author        =     {Ruijie Lu and Yu Liu and Jiaxiang Tang and Junfeng Ni and Yuxiang Wang and Diwen Wan and Gang Zeng and Yixin Chen and Siyuan Huang},
	title            =     {Generating Objects with Part-Articulation from a Single Image},
	booktitle    =    SIGGRAPHAsia,
	pages         =    {155:1--155:13},
	year            =    {2025},
}

@inproceedings{Zhao-2026-Functionalization,
	author       =      {Mingrui Zhao and Sai Raj Kishore Perla and Kai Wang and Sauradip Nag and Duc Anh Nguyen and Jiayi Peng and Ruiqi Wang and Angel X. Chang and Manolis Savva and Ali Mahdavi-Amiri and Hao Zhang},
	title            =      {Functionalization via Structure Completion and Motion Rectification},
	booktitle   =      SIGGRAPHAsia,
	year           =      {2026}
}

@inproceedings{Arabi-2026-EUREXA,
	author      =      {Abul Al Arabi and Charles Yushi Cai and Ryann Lu and Shu Kong and Jeeeun Kim},
	title           =      {{EUREXA}: End-User Reconfiguration of Environment with e{X}plainable Augmentation for Generative Fabrication},
	booktitle   =     CHI,
	pages         =    {799:1--799:17},
	year         =      {2026}
}

@inproceedings{Guo-2025-CRAFT,
	author       =    {Michelle Guo and Mia Tang and Hannah Cha and Ruohan Zhang and C. Karen Liu and Jiajun Wu},
	title            =    {{CRAFT}: Designing Creative and Functional {3D} Objects},
	booktitle    =    {Proc. of the IEEE/CVF Winter Conference on Applications of Computer Vision (WACV)},
	pages        =    {7215--7224},
	year           =    {2025},
}
\end{document}